\documentclass[review]{elsarticle}

\usepackage{lineno,hyperref}
\usepackage{amsmath}
\usepackage{mathtools}
\usepackage{amsfonts}
\usepackage{subcaption}
\usepackage{caption}
\usepackage{tikz}
\usepackage{MnSymbol}%
\usepackage{wasysym}%
\usepackage{lscape}
\usepackage{pdflscape}
\usepackage{chemformula} 
\usepackage{siunitx}
\usepackage{kotex} 

\modulolinenumbers[5]

\newcommand{\veta}{\boldsymbol{\eta}}

\newcommand{\figref}[1]{Fig. \,\ref{#1}}

\graphicspath{{Figs/}} 

\newcommand{\rom}[1]{\uppercase\expandafter{\romannumeral #1\relax}}

\begin{document}

\captionsetup[figure]{labelfont={bf},labelformat={default},labelsep=period,name={Fig.}}
\captionsetup[table]{labelfont={bf},labelformat={default},labelsep=newline,name={Table}, justification=raggedright, singlelinecheck=off}

\begin{frontmatter}

\title{Optimal Planning of Hybrid Energy Storage Systems using Curtailed Renewable Energy through Deep Reinforcement Learning}

\author[SNU]{Dongju Kang\corref{1st}}
\author[EWHA,BK21]{Doeun Kang\corref{1st}}
\author[EWHA,BK21]{Sumin Hwangbo}
\author[PNU]{Haider Niaz}
\author[SNU]{Won Bo Lee}
\author[PNU]{J. Jay Liu}
\author[EWHA,BK21]{Jonggeol Na \corref{mycorrespondingauthor}}
\ead{jgna@ewha.ac.kr}

\cortext[mycorrespondingauthor]{Corresponding author}
\cortext[1st]{The two authors have the same contribution to this study}

\address[SNU]{School of Chemical and Biological Engineering, Seoul National University, Gwanak-ro 1, Gwanak-gu, Seoul, 08826, Republic of Korea.}
\address[EWHA]{Department of Chemical Engineering and Materials Science, Ewha Womans University, Seoul 03760, Republic of Korea}
\address[BK21]{Graduate Program in System Health Science and Engineering, Ewha Womans University, Seoul 03760, Republic of Korea}
\address[PNU]{Department of Chemical Engineering, Pukyong National University, Busan 48513, Republic of Korea}

\begin{abstract}
Energy management systems (EMS) are becoming increasingly important in order to utilize the continuously growing curtailed renewable energy. Promising energy storage systems (ESS), such as batteries and green hydrogen should be employed to maximize the efficiency of energy stakeholders. However, optimal decision-making, i.e., planning the leveraging between different strategies, is confronted with the complexity and uncertainties of large-scale problems. Here, we propose a sophisticated deep reinforcement learning (DRL) methodology with a policy-based algorithm to realize the real-time optimal ESS planning under the curtailed renewable energy uncertainty. A quantitative performance comparison proved that the DRL agent outperforms the scenario-based stochastic optimization (SO) algorithm, even with a wide action and observation space. Owing to the uncertainty rejection capability of the DRL, we could confirm a robust performance, under a large uncertainty of the curtailed renewable energy, with a maximizing net profit and stable system. Action-mapping was performed for visually assessing the action taken by the DRL agent according to the state. The corresponding results confirmed that the DRL agent learns the way like what a human expert would do, suggesting reliable application of the proposed methodology.
\end{abstract}


\end{frontmatter}

\newpage
\tableofcontents
\newpage

\section{Introduction}\label{sec:intro}

Renewable energy sources such as biomass, hydro-power, wind, and solar power, could become alternative energy sources for the future and reduce humankind's dependence on fossil fuels \cite{DINCER2000157}. Installation of solar panels and wind generators has exponentially increased owing to the renewable portfolio standards, which area regulations aimed at increasing the energy production from renewable sources \cite{GOLDEN201536}. The implementation of these regulations, have significantly increased the penetration rate of renewable energy power, and consequently, the problem of oversupply has merged \cite{IMPRAM2020100539}, causing a disproportion between demand and supply for each period. Thus, energy curtailment, i.e., a deliberate reduction in the output to levels below the original power generation capacity, has been implemented to control excess power generation \cite{9224611, BIRD2016577}; however, the energy is still scarce during peak hours (\figref{fig:year_energy}(a) and (b)). Therefore, it would be economically and environmentally profitable to integrate the curtailed energy into energy storage systems (ESS) rather than installing more power generators \cite{SABER2019414, 8125737}.

Traditionally, this type of energy management planning problem is solved using mathematical programming (MP). MP is a type of optimization used to maximize or minimize objective function from a mathematical model that exhibits features and characteristics of some other problems with a set of mathematical relationships (such as equations and inequalities) \cite{williams2013model}. However, renewable energy resources, such as wind and solar power, fluctuate significantly depending on the daily weather and environmental conditions; therefore, regular power generation cannot be guaranteed \cite{LI20151067}. The curtailed renewable energy, which is the excess electricity generated from renewable sources, contains the uncertainties in both generation and demand, resulting in a significant uncertainty, and difficult to forecast \cite{6936947}. Therefore, integrating curtailed energy into an ESS causes random variations in both supply and demand \cite{6345450}. In the field of MP, there are some methods like stochastic optimization (SO) and robust optimization (RO) that can optimize the objective function while considering the uncertainty. However, these methods require enormous computation power as the number of variables, and thus the uncertainty, increases. Therefore, it is difficult to expand and apply these methods to a large and complex problem \cite{8769895}, indicating that renewable power curtailment planning with MP is stymied by several limitations.

Reinforcement learning (RL) has emerged as an alternative method that makes up for MP and solves large and complex problems \cite{DREHER2022115401,ZHANG2019112199}. RL is formalized by using the optimal control of incompletely-known Markov decision processes and learning what action to take in a given situation \cite{sutton2018reinforcement}. Among the many RL algorithms, policy-based RL works more effectively than value-based RL in systems with uncertainties, because it expresses the policy itself as a probability\cite{ZHANG2021113608,ZHANG2021114381}; therefore, unlike conventional methods that converge to one optimal action, policy gradient algorithms could learn stochastic policy \cite{pmlr-v32-silver14}. However, the existing RL algorithms had limitation of lacked scalability because of the complexity problems. The state-of-the-art deep reinforcement learning (DRL), contains the function expressed as a deep neural network, which makes it possible to learn more complex associations concerning variable input values \cite{8103164}. Therefore, it does not take up much computational power and memory, even if the size and the complexity of the problem increases, and the curse of the dimension of the NP-hard problem can also be avoided \cite{en12122291}. In addition, DRL can have a scalable solution by training an agent in a specific environment in the first place, and it can be applied to other cases and scenarios \cite{8331897}, such the training time can be greatly reduced.

For these reasons, the application of RL in solving the energy management problem is being actively studied. Owing to the complexity of the energy management problem, conventional methods appear to be inapplicable. However, a DRL algorithm is applied to obtain the desired control scheme, and a practical solution is obtained despite the stochastic and uncertain nature of the power output \cite{HUA2019598,doi:10.1080/17445760.2014.974600,8742669}. The double deep Q-learning method, an DRL method, possesses the uncertainty rejection capability through the utilization of a deep neural network as a function approximator \cite{8742669}. The DRL algorithm proposed by B Zhang. et al. \cite{ZHANG2020113063}, showed that can increase the profit of the system operator, reduce wind power curtailment in real time in an integrated electricity and natural gas system with renewable energy. However, studies on quantitative analyses with MP-based methodologies, such as SO or RO, have not been conducted in the field of energy management system thus far. Since the rationality of the trained agent's decision could not be interpreted based on human intuition, it could be said to be insufficient in terms of interpretability. Further, a methodology to build the DRL model in a form more suitable to the problem and analyze the corresponding results according to the uncertainty is required.

Here, we propose an DRL-based real-time planning artificial intelligence that decides how to store the curtailed energy generated in a micro-grid into a battery energy storage system (BESS) or green hydrogen produced by an alkaline water electrolyzer (AWE). This allows flexibility and a reasonable performance in our problem that has hybrid inputs and multiple outputs by using policy-based agents. Further, we quantitatively verify the real-time decision performance of the DRL by comparing the trained policy network with the SO that knows the variable values of all the episodes. Next, the trained policy is shown in the action mapping figure that can be intuitively accepted by humans, through mapping and schematizing the action according to the state. Analysis through this action mapping figure enables humans to logically accept and understand the outcome of the policy. Finally, in order to confirm the accuracy of the specific value of the DRL, a part of the actual data and the DRL components (such as action, environment, and state) are shown to reflect how well the DRL trained.

\section{Problem Description}\label{sec:exp}

\subsection{Structure of the Hybrid Energy System}
\begin{figure}
	\includegraphics[width=\textwidth]{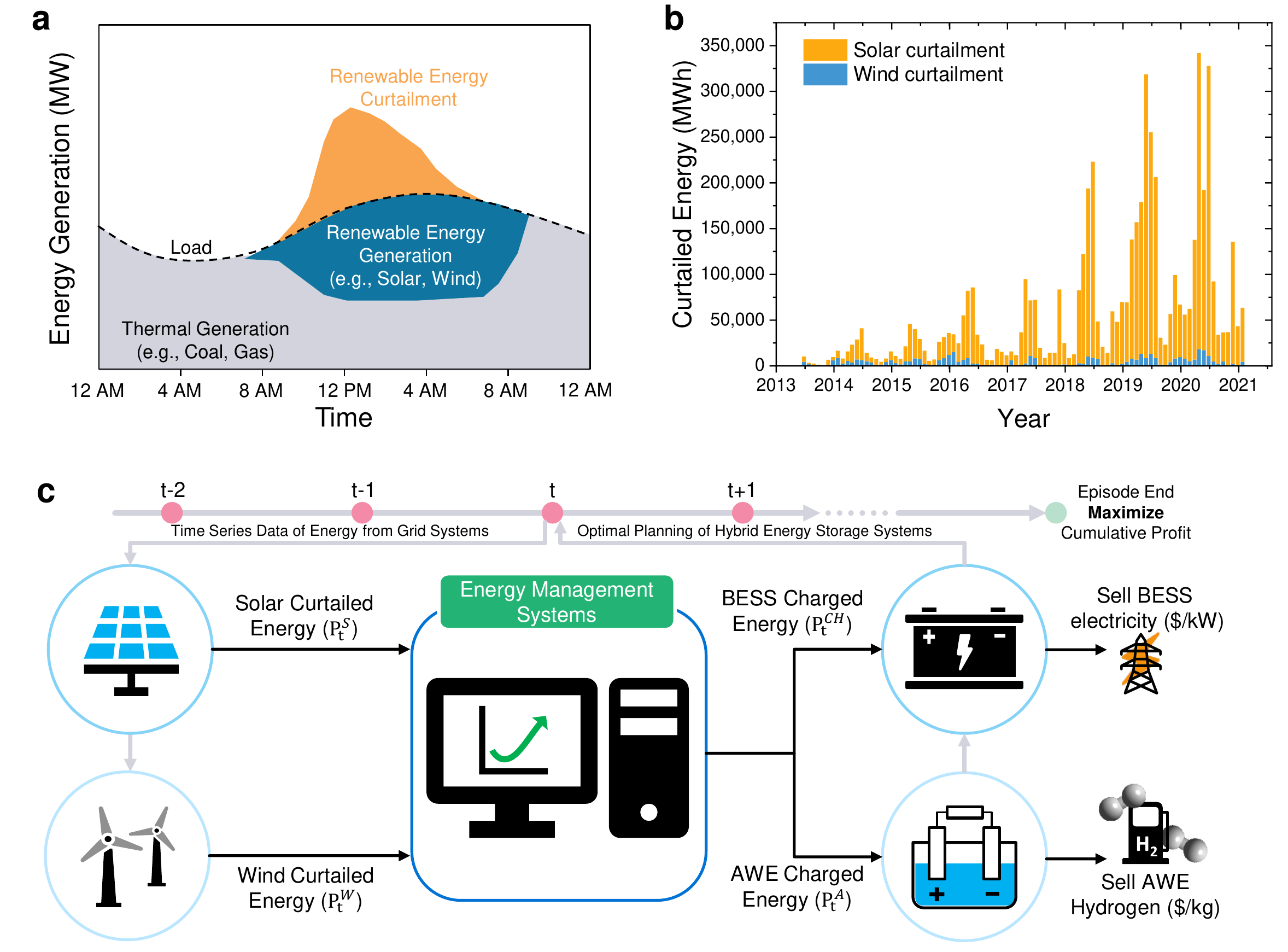}
	\caption{Optimal planning of ESSs with the renewable energy curtailment. (a) Schematic illustration of renewable energy curtailment. (b) Solar and wind energy curtailment in California from 2013 to 2021. Data adapted from ref. \cite{CAISO_data}. (c) Schematic diagram of the optimal planning problem of hybrid energy storage systems covered in this study.}
	\label{fig:year_energy}
\end{figure}

For the curtailed renewable energy utilizing system, we used two renewable energy sources, wind and solar, in California \cite{CAISO_data}. As the two energy sources become one and enter the energy system, a new type of pattern is revealed. There are two ways for utilizing the curtailed energy: a BESS and AWE. Information on the curtailed energy generated every hour is transferred to the Energy Management System (EMS), and the energy is stored through ESS, a BESS or AWE, according to the internally calculated planning decision-making. The ultimate goal of this optimal planning problem is to make a decision in a direction that maximizes the profit obtained during the entire operating period (e.g., one year) (\figref{fig:year_energy} (c)). From a usage point of view, BESS and AWE have different characteristics, and both are viable alternatives. BESS is a type of storage device with incompatible charging and discharging, which is one of the reasons why we need to make an optimal decision. Charging and discharging are depended on the state of charge (SOC) of BESS, and this issue leads to the BESS sizing problem while deciding the battery capacity. In contrast, the AWE is a type of conversion device with no incompatible actions, because it has only the options of turning on and off. The energy form is converted into hydrogen, which does not require a special capacity of the AWE to store energy like SOC. AWE is useful when the curtailed renewable energy exceeds the initial or residual SOC of the BESS. The dynamic constraints of BESS are more complicated than those of the AWE, but the electricity is more cost-valuable compared to hydrogen. It can be changed in areas where the electricity price is less than half of those used in this study. Therefore, it is more effective to combine the two utilization methods in management systems. Further, the optimal decision-making model needs to choose not only which utilization system would be used but also how much energy should be distributes between the BESS and AWE at every unit time $t$.

\subsection{Mathematical Model}\label{sec:full}
The deterministic mathematical model is used to find the optimal solution of the problem. The aim of this study was to maximize an objective function value, which is a profit of the ESS. The specific data on the curtailed renewable energy in 2020 in California (detail in section 2.3) were used for the deterministic mathematical model, which is a mixed-integer linear programming (MILP) and CPLEX 20.1.0.1 was used as the solver. Preliminary mathematical study of the system can be found in ref. \cite{SHAMS2021}.

	\begin{table}[t]
		\caption{Notation}
		\label{tbl:notation}
		\centering
		\resizebox{\textwidth}{!}{%
		\begin{tabular}{@{\extracolsep{\fill}}cl}
			\hline
			\textbf{Indices}  &   \\
			$t$        &     Index of time, t=1,2,3...T  \\
			\hline 
			\textbf{Parameters}   & \\ 
			$\eta^{CH}$ &   Charging efficiency \\
			$\eta^{DH}$ &   Discharging efficiency  \\
			$\eta^{AWE}$ &  AWE efficiency \\
			$\alpha$    &   Co-efficient for BESS capacity \\
			$\beta$ &   Co-efficient for AWE capacity \\
			$LHV$   &   Lower heating value of hydrogen \\
			
			\hline
			\textbf{Binary Variables}   &   \\
			$z^{CH}_{t}$ &   Charging on/off at time $t$ \\
			$z^{DH}_{t}$ &   Discharging on/off at time $t$ \\
			$z^{AWE}_{t}$   &   AWE on/off at time $t$ \\
			
			\hline
			\textbf{Variables}   &   \\
			$P^{S}$  &   Solar curtailed power at time $t$ \\
			$P^{W}$  &   Wind curtailed power at time $t$ \\
			$P_{t}^{CH}$  &   BESS charging at time $t$ \\
			$P_{t}^{DH}$  &   BESS discharging at time $t$ \\
			$P_{t}^{A}$   &   Power to AWE at time $t$\\
			$CF_{t}$  &   Curtailed power fraction at time $t$ \\
			$AWE^{P}$   &   AWE power\\
			$BESS^{P}$  &   BESS power\\
			$H_{t}^{AWE}$  &   Amount of hydrogen produced at time $t$\\ 
			$REV^{BESS}$, $REV^{AWE}$   &   Revenue from BESS and AWE\\
			$S^{DH}$, $S^{H_{2}}$    &   Sale price of discharge and hydrogen power\\
			$C^{O\&M}$ &   Operations and maintenance cost\\
			$C^{ACC}$ &   Annualized capital cost\\
			$C^{VOM}$ &   Variable operations and maintenance cost\\
			$C^{FO}$ &   Fixed operations cost\\
			\hline
		\end{tabular}
        }
	\end{table}

First, at time $t$, the curtailed wind ($P^{W}_{t}$) and solar ($P^{S}_{t}$) energy is combined and used as the energy charged in the BESS ($P^{CH}_{t}$) and that converted to hydrogen in the AWE ($P^{A}_{t}$) can be expressed as the following equation:
\begin{eqnarray}
CF_{t} \times (P^{W}_{t}+P^{S}_{t})=P^{A}_{t}+P^{CH}_{t}\label{eq:used curtailed energy}\\
0 \leq CF_{t} \leq 1\label{eq:curtailed fraction}
\end{eqnarray}
where $CF_{t}$ is the curtailed fraction at time $t$.

For the BESS constraints, two binary variables are required, which represent the status of charging ($z^{CH}_{t}$) and discharging ($z^{DH}_{t}$) at time $t$ since the two status cannot occur simultaneously:
\begin{eqnarray}
z^{CH}_{t} + z^{DH}_{t} \leq 1 \label{eq:constraint of charging and discharging}
\end{eqnarray}

The amount of BESS charging and discharging at time $t$ is determined by each of the charging and discharging on/off binary variables and the BESS minimum/maximum power ($BESS^{P, MIN}$, $BESS^{P, MAX}$):
\begin{eqnarray}
z^{CH}_{t} \times BESS^{P, MIN} \leq P^{CH}_{t} \leq z^{CH}_{t} \times BESS^{P, MAX}\label{eq:charge constraint}\\
z^{DH}_{t} \times BESS^{P, MIN} \leq P^{DH}_{t} \leq z^{DH}_{t} \times BESS^{P, MAX}\label{eq:discharge constraint}
\end{eqnarray}
Further, the BESS power is set to 30 \% of the BESS capacity to prevent BESS charging overload \cite{kim2020active}.

Because the amount of energy that can be stored in the BESS is limited depending on its capacity, there is an SOC constraint:,
\begin{eqnarray}
SOC_{t} = SOC_{t-1} + P^{CH}_{t} \times \veta^{CH} - P^{DH}_{t}/\veta^{DH},  t>1\label{eq:state of charge}\\
\alpha \times BESS^{CAP} \leq SOC_{t} \leq BESS^{CAP}\label{eq:soc constraint}
\end{eqnarray}
where $\veta^{CH}$ and $\veta^{DH}$ are the charging and discharging efficiencies, respectively, and $\alpha$ is the co-efficient of BESS capacity.

For the AWE constraints, the AWE power at a time $t$ is determined as
\begin{eqnarray}
z^{AWE}_{t} \times AWE^{P, MIN} \leq P^{A}_{t} \leq z^{AWE}_{t} \times AWE^{P, MAX}\label{eq:awe constraint}
\end{eqnarray}
and the AWE minimum power ($AWE^{P, MIN}$) is decided by the AWE maximum power ($AWE^{P, MAX}$) as
\begin{eqnarray}
AWE^{P, MIN} = \beta \times AWE^{P, MAX}\label{eq:awe min}
\end{eqnarray}
where $\beta$ is the coefficient of the AWE capacity.
The amount of hydrogen produced at time $t$($H^{AWE}_{t}$) can be calculated in kg unit as follows:
\begin{eqnarray}
H^{AWE}_{t} = P^{A}_{t} \times (\eta^{AWE}/LHV)\label{eq:hydrogen produced}
\end{eqnarray}
where $\eta^{AWE}$ is the AWE efficiency, and $LHV$ is the lower heating value of hydrogen.

The cumulative of the discharged power ($T^{{P}^{DH}_{t}}$) and produced hydrogen ($T^{{H}^{AWE}_{t}}$) during a period can be summed as,
\begin{eqnarray}
REV^{BESS} = \sum_{t=1}^{T}P^{DH}_{t} \times S^{DH} \label{eq:electricity revenue} \\
REV^{AWE} = \sum_{t=1}^{T}H^{AWE}_{t} \times S^{H_{2}} \label{eq:hydrogen revenue}
\end{eqnarray}
where $T$ is the total length of the period. These values directly affect the calculation of the profit during a period.

The objective function that represents the net profit through the operation of this hybrid energy system can be defined as
\begin{eqnarray}
OBJ =& (REV^{BESS}+REV^{AWE}) \nonumber \\ 
     & -(C^{O\&M, BESS}+C^{ACC, BESS}+C^{O\&M, AWE}+C^{ACC, AWE})
\label{eq:objective function}
\end{eqnarray}

where $REV^{BESS}$ and $REV^{AWE}$ are the revenues generated from the BESS and AWE, respectively; $C^{O\&M}$ and $C^{ACC}$ are the operations and maintenance cost and the annualized capital cost, respectively. The detailed description of the cost estimation process is described in Supplementary Note 5.

\subsection{Datasets}
The data used in this study are the actual curtailed energy data for California, provided by the California Independent System Operator \cite{CAISO_data}. The raw data are divided into wind and solar curtailed energy data with 15-min intervals. For a higher efficiency, the data was processed into a total of 8760 h or 8784 h (leap year) for one year, summed in units of one hour. Since the renewable energy generation facilities have grown exponentially, only the data from 2019 to 2021 were used to reflect the latest trends. For the product price, the electricity price was taken from the hourly data of the Southern California Edison Time-of-Use rate plans, \cite{Electricity_price} which reflect the energy demand peak per day, and the hydrogen price was set to a constant value of \$6/kg \cite{hydrogen_cost}.

\section{RL Architecture}\label{sec:full}
RL refers to learning a policy to identify what actions to take depending on the situation to maximize the numerical reward signal \cite{sutton2018reinforcement}. The agent, which is the subject of the learning, receives information about the state and updates the policy by receiving a reward after taking an action. This process proceeds sequentially and is based on a trial-and-error search method. To build our RL model, we constructed the environment that describes the dynamics of the hybrid energy system, and designed the state and reward. To apply the RL algorithm in connection with the environment, RAY RLlib version 1.9.0. \cite{moritz2018ray, RAY_RLlib}, which is an open source library, was used.

\subsection{Neural-Network Architecture and Learning Loop}
\begin{figure}
	\includegraphics[width=\textwidth]{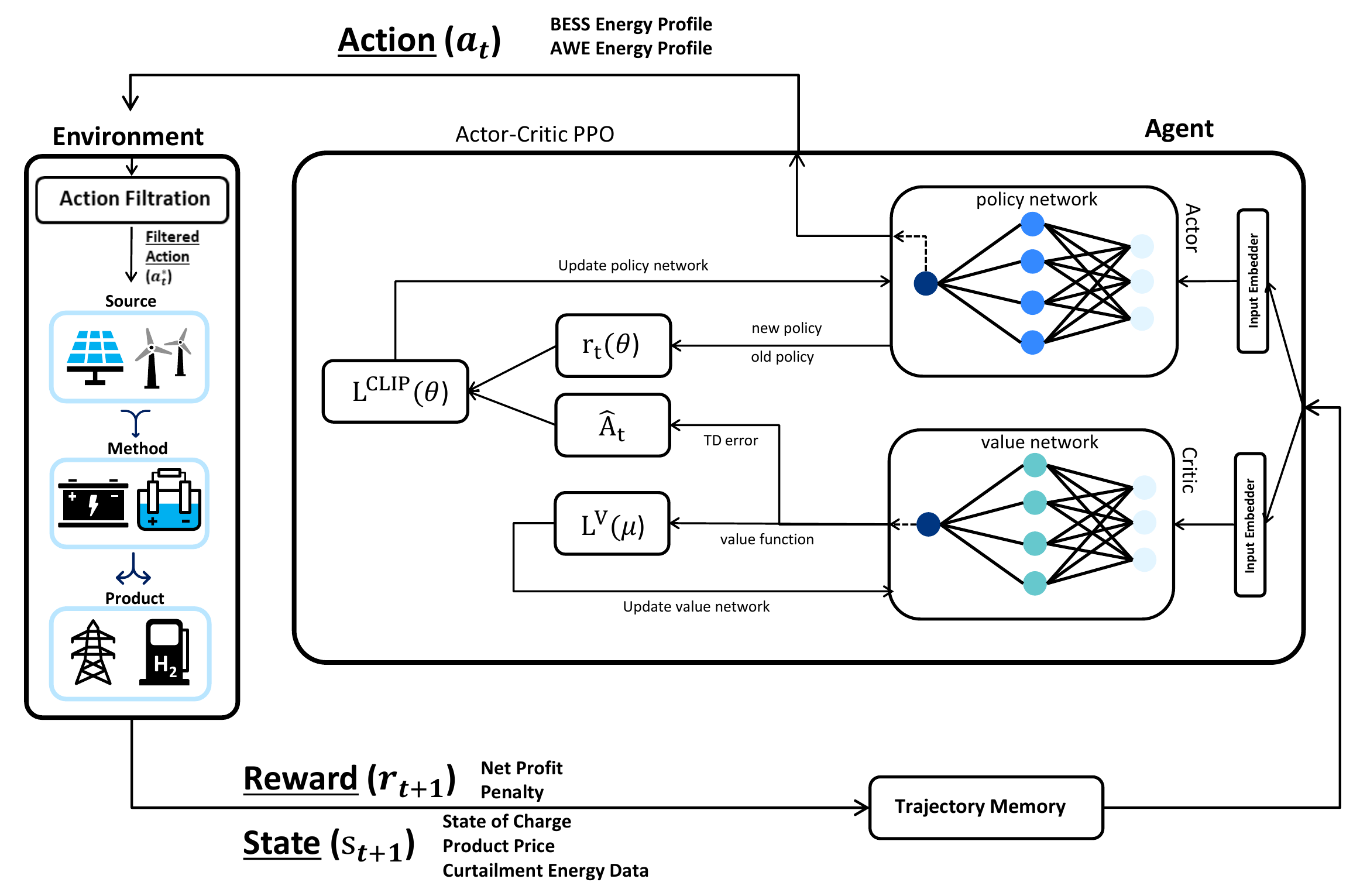}
	\caption{Representation of the components of the proposed DRL-based real-time planning of the hybrid energy system architecture.}
	\label{fig:RLstructure}
\end{figure}

In all the DRL algorithm training used in this study, the Proximal Policy Optimization (PPO) algorithm was selected because of its high performance, stability and training speed compared to that of the other algorithms in our environment \cite{https://doi.org/10.48550/arxiv.1707.06347}. This algorithm was developed based on the policy gradient algorithm, and trains a policy network consisting of the probability distribution of actions taken in a specific state. Thus, if both the state and action space are large, such as the problem solved in this study, the PPO is likely to exhibit a performance advantage by adopting a stochastic policy. In other words, the policy is expressed using a neural-network called the policy network. We organized the neural-network with three fully connected layers and each layer has 256 hidden nodes. $tanh$ is used as the activation function, and $relu$ is used as the activation function for the final fully connected stack of dense layer. \figref{fig:RLstructure} shows how the PPO algorithm works in our environment. For the other hyper-parameter, the learning rate and train batch sizes are set to $3 \times 10^{-5}$ and 8000, respectively. These values are tuned with the \textit{HyperOptSearch} algorithm which is based on the Tree-Parzen Estimators \cite{HyperOpt}.


	
\subsection{Environment, State, and Action}
The environment includes everything except the agent such as the curtailed power data, SOC, production price, and so on. State expresses how the environment is changed by the actions, and the agent gets the information of the state by observation. In this case, the agent observes the production prices, previous SOCs, and amount of curtailed energy. The agent state was designed in two forms depending on how it receives information on the curtailed energy; these states were named as internal prediction (IP) and external prediction (EP). The configuration of the IP state contains the past data of the curtailed energy; therefore, the agent uses the data to predict the amount of curtailed energy at a particular moment of the state to make a decision. In the EP state, we assume an external model that predicts the curtailed energy, and the EP state receives the predicted values as the state.
In this system, the agent decides to make a decision about how much to charge or discharge the battery as well as how much of the curtailed energy should be sent to the AWE. These deciding variables correspond to actions in RL and require a continuous range of action spaces. We set the BESS charging and discharging in one continuous action that changes the SOC state to "plus" or "minus". The "plus" and "minus" signs of the BESS action imply charging and discharging, respectively. In AWE, the action is taken from 20 \% of the maximum capacity to the maximum capacity, because the AWE is not activated when the power is less than 20 \% of the AWE power capacity. The action sometimes produces an invalid value, which is called overaction and can be divided into over-charging and over-discharging. Charging higher than the remaining SOC is called over-charging, whereas discharging higher than the SOC is called over-discharging. Preventing these overactions is crucial to avoid fire hazard and prolong the battery life.

\subsection{Rewards and action filtration}
The objective function of the MP, i.e., net profit, is modeled as the reward in RL. The crucial difference in expressing the objective function of the MP and RL is that the agent of RL receives the reward at every time step, whereas there is only one value of the objective function in MP. In this case, the episode length ($T$) is the number of hours of the training curtailed data (i.e. 8784 h in the year 2020). Therefore, the elements included in CAPEX of the objective function are manipulated by scaling in units of hour and then applied as rewards at time $t$ ($\mathcal{R}_t$).
\begin{eqnarray}
C^{O\&M} = C^{VOM} + C^{FO}\label{eq:O&Mcost} \\
\mathcal{R}_t = REV_{t}-C^{VOM}_{t}-(C^{FO}+C^{ACC})/T \label{eq:reward_nopenalty}
\end{eqnarray}
where $C^{VOM}$ is the variable operation and management cost; $C^{FO}$ is the fixed operation and management cost; $REV$ and $C^{VOM}$ are changed to terms by time step $t$; $C^{FO}$ and $C^{ACC}$ are divided by $T$ to be applied to only one time step.
 
Our aim is to make decisions on raising profits from the energy system operation as well as to ensuring that the system can be operated safely and for a long time. Therefore, we also imposed penalties when the decisions made by the agent lead to over-charging or over-discharging in the BESS. Since the direction of pursuit of the operation method may be different, the method of granting the penalty can vary.

\begin{equation}
  \mathcal{P}_t =
  \begin{cases}
    P^{CH}_{t}-(100-SOC_{t})\times BESS^{cap} & \text{if $P^{CH}_{t} \geq (100-SOC_{t}) \times BESS^{cap} $} \\
    P^{DH}_{t}-(SOC_{t}-SOC_{min})\times BESS^{cap} & \text{if $P^{DH}_{t} \geq (SOC_{t}-SOC_{min})\times BESS^{cap}$} \\
    0 & \text{otherwise}
  \end{cases}
\end{equation}
where $BESS^{cap}$ is the BESS capacity, i.e., 1500MWh.

Therefore, the final reward can be expressed as follows.
\begin{eqnarray}
\mathcal{R}_t = REV_{t}-C^{VOM}_{t}-(C^{FO}+C^{ACC})/T-\mathcal{P}_t \times \mathcal{F} \label{eq:reward_penalty}
\end{eqnarray}
where $\mathcal{P}_t$ and $\mathcal{F}$ are the penalty size at time $t$ and the penalty scaling factor, respectively.

However, imposing penalty is not always sufficient to eliminate all the overactions. To avoid the invalid value of the action while maintaining a high profit, action filtration is also configured in the environment. The action filtration is a system used to prevent the agent from making overactions due to an error, even when the agent makes a decision according to the flow of the incoming curtailed energy. If the action is over-charging, then it is reduced such that the SOC does not exceed 100 \%, and in the case of over-discharging, discharging only up to a certain range, except for the minimum capacity, is allowed.

\section{Results \& Discussion}\label{sec:exp}
\subsection{DRL Performance}\label{sec:full}

\begin{figure}
	\includegraphics[width=\textwidth]{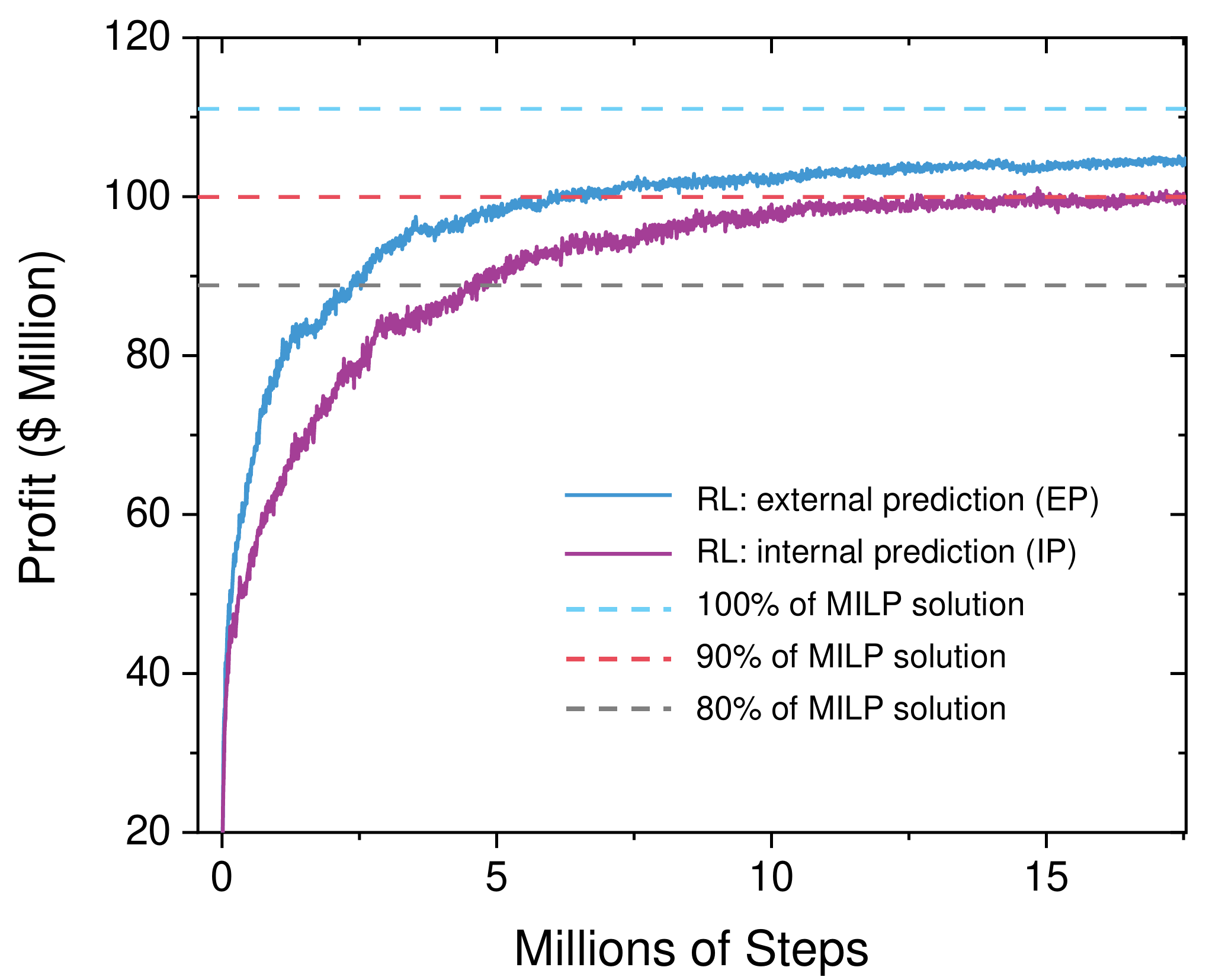}
	\caption{Profit also rises as the DRL training progresses. The three horizontal lines indicates a relative value based on the MP optimal value. The profit of the EP and the IP are 1.05$\times 10^8$ and 1.00$\times 10^8$, respectively, whereas the MP's optimal value is 1.11$\times 10^8$.}
	\label{fig:training}
\end{figure}

In both the states, EP and IP, the agent takes a random action in the beginning and shows a very low profit. As the training progresses, the profit rapidly rises and then converges. Since the reward is a value for one step, it is necessary to compare the result through the sum of the rewards for one episode (i.e, 8784 steps), i.e., the return.  However, the reward contains a penalty that does not exist in the MP model, making it difficult to perform an accurate comparison with the MILP solution. By calculating the profit value from the action during training, it can be directly confirmed that both the EP and IP show a high performance (over 90 \%) compared to the optimal value which is the deterministic MILP solution (\figref{fig:training}). The maximum value of the accumulated reward without penalty during one episode could be equal to the solution of the objective function of MILP. To be more precise, EP and IP show a profit of almost 94 \% and 90 \%, respectively, compared to MILP solution. This is a high performance for what is achieved with a minimal hyperparameter tuning and making decisions in real time. Moreover, in the case of IP, a conspicuous performance was observed even though there was no model to predict the curtailed renewable energy. Because an additional curtailed energy prediction model is not required, IP can be considered to be an efficient model in terms of time and computational cost.

As a result of designing rewards in various ways to improve the DRL training, it was confirmed that the characteristics of the policy and the final reward value vary according to the penalty scale (Fig. S1). In particular, if you have an IP state, then the effect of the penalty scale is greater. The absence of predicted curtailed energy values indicates the importance of predicting the curtailed energy. However, if the penalty scale becomes larger than a certain level, then the agent training focuses too much on avoiding the penalty rather than increasing the profits, thereby decreasing the profits. We found a section where overaction does not occur too much, and a high reward is maintained. Further, the training was conducted using a scale: 100 times the normal charged and discharged energy values. The detailed results according to the penalty scale are described in the Supplementary Note 1. The data for one year in 2020 are used for the training, and we applied this data in two ways: deterministic data using the values of the original data without making any changes, and stochastic data, in which the data exhibit fluctuations and are trained in every episode during the training. However, it was judged that learning various patterns of the training data is more important than introducing variability to the training data (See in Table S1 and Fig. S2). Therefore, only the deterministic training data were used to obtain the results in this study. In this way, the DRL model was trained over 1.5$\times 10^7$ steps as shown in \figref{fig:training}.


We further manipulated the length of the data period used as the training data and the period of information received as the state to investigate the factors that influence the DRL training. As a result, we decided that the most appropriate period is one year for the training data and receive 24 h of information as the state. The above training uses these values and the training results for other values are described in detail in Fig. S3 and S4.

\subsection{Policy evaluation under uncertainty}\label{sec:full}

\begin{figure}
	\includegraphics[width=\textwidth]{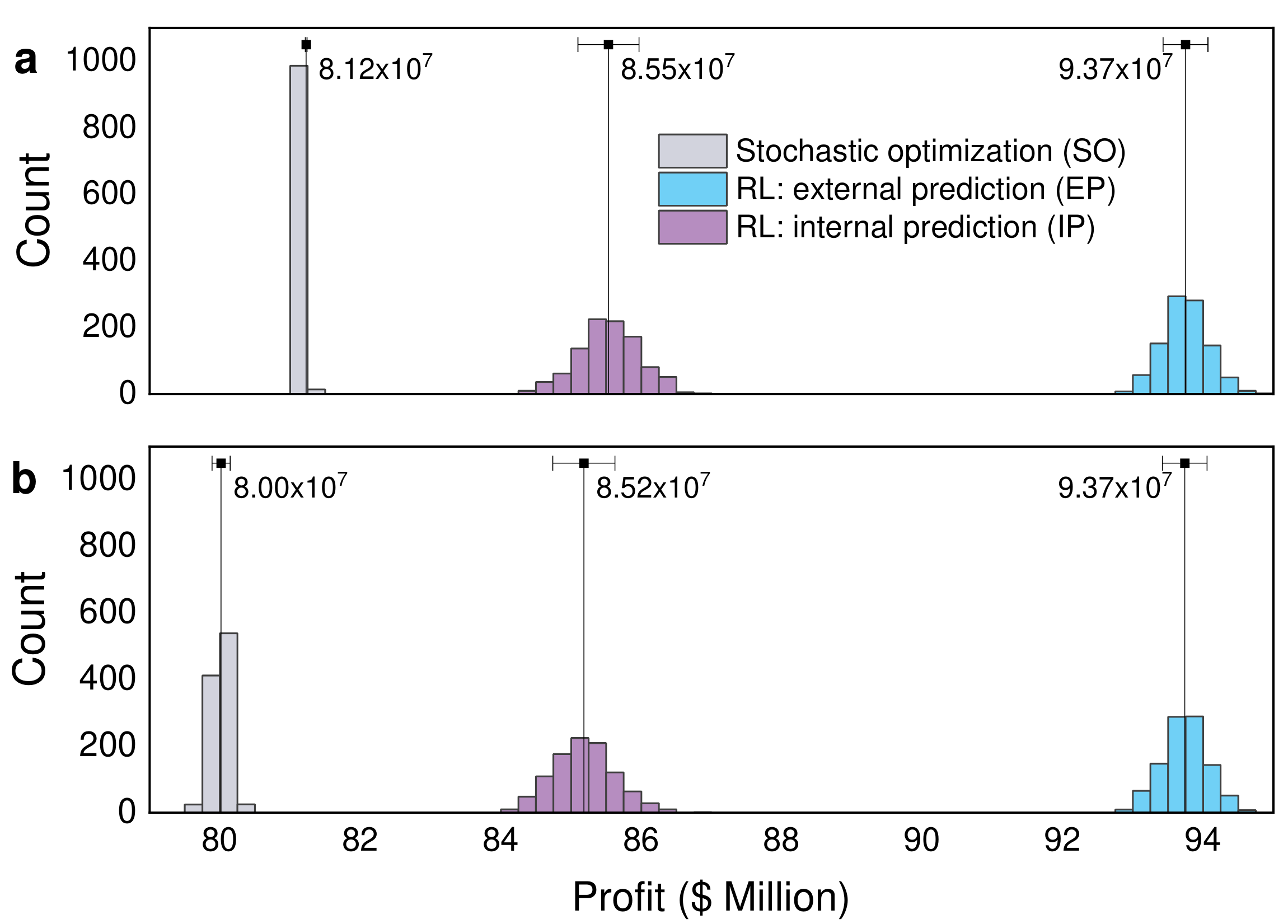}
	\caption{Monte Carlo simulation results with diverse curtailed energy scenarios. The same 1000 scenarios, which are manipulated from 2021 data, were used for the SO, IP and EP. The scenario with $\pm$10 \% and $\pm$20 \% uncertainty in the incoming curtailed energy, are used for each graph (a) and (b), respectively.}
	\label{fig:histogram}
\end{figure}

All the policies trained thus far show good rewards near the deterministic MILP solution. Compared to the SO, which is an MP method considering the uncertainties, DRL shows better performance when there is an uncertainty in the curtailed energy, as proven through Monte Carlo simulations (\figref{fig:histogram}). Thus, DRL is a powerful method against the uncertainty of the incoming curtailed energy. Although the DRL agent with the EP state generates higher profits than that with the IP state, the agent with the IP state can still achieve much higher profits than that of the SO. Moreover, SO is not a real-time decision-making method like DRL. It is a plan obtained by giving all information for the year 2021. Therefore, it can be said that the overall performance of DRL is higher than that of the SO. If we change the magnitude of the considered uncertainty of the scenarios from $\pm$10 \% to $\pm$20 \%, then the strength of the uncertainty of the DRL becomes more pronounced. As the uncertainty increases, the profit drop of the SO is larger than that of the DRL (i.e., the RL shows no significant the profit drop).

\begin{figure}
	\includegraphics[width=\textwidth]{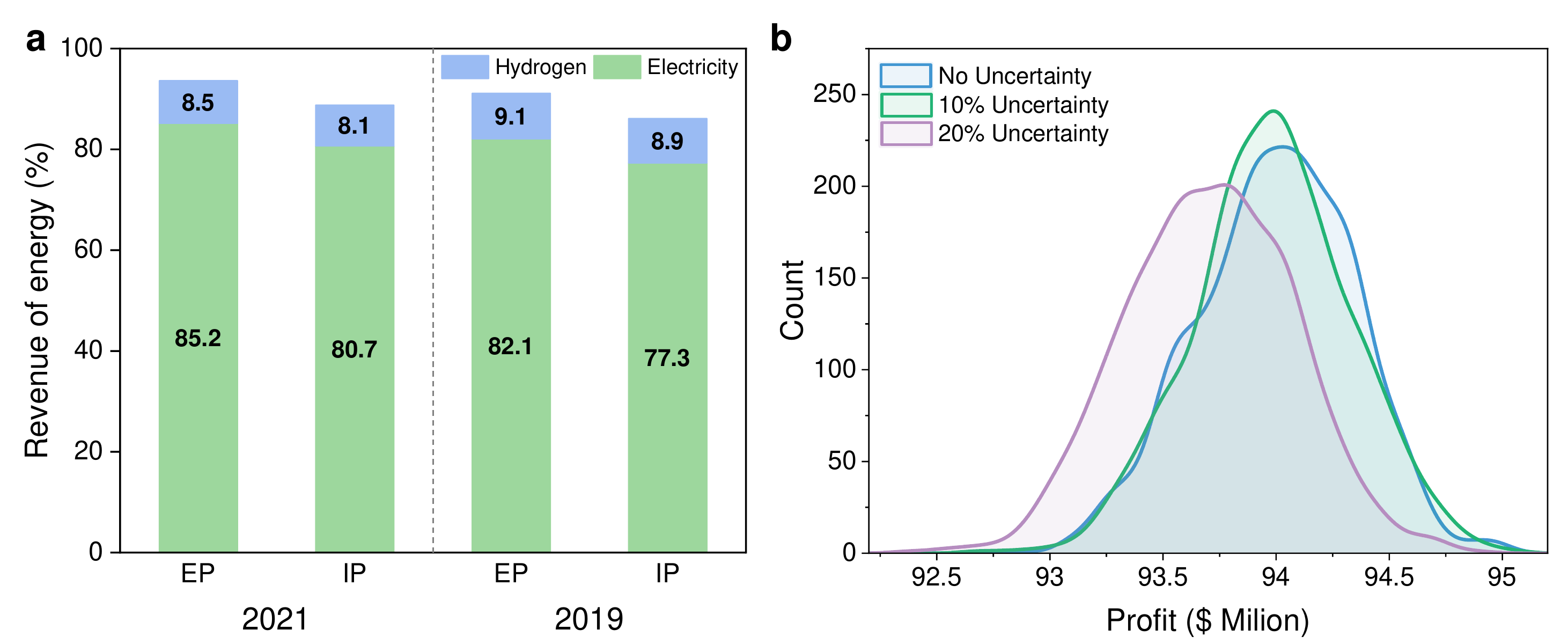}
	\caption{Results with various inputs for the DRL model to verify the DRL policy. (a) Relative profits against the optimal value that are evaluated with 2019 and 2021 data. Profits from hydrogen and electricity are shown separately. (b) Monte Carlo evaluation of how the EP model performs when there are errors in the predicted value of the curtailed energy. The same EP model was used for all the three distribution of the evaluation results with different curtailed energy accuracies.}
	\label{fig:profit&distribution}
\end{figure}

To confirm the more intuitive result, the evaluation result was compared with the optimal value, MILP solution, through deterministic scenarios. For the scenarios, the real curtailed renewable energy data of 2021 and 2019 were used for evaluation. Although the agents were trained with the data of 2020, the profits in the 2019 and 2021 scenarios remained high between the late 80 \% and early 90 \% relative to the optimal values for each year (\figref{fig:profit&distribution} (a)). This result also can be interpreted as the DRL agent was not over-fitted by the training data.

The EP state DRL model continues to show slightly better results than the IP state one, which is not realistically possible because the prediction values used in the EP state are assumed to be 100 \% accurate. However, based on the Monte Carlo simulation result, it was confirmed that the error in the predicted curtailed energy value did not significantly affect the performance of the agent with the EP state. Curtailed energy information with a low accuracy was used in the EP model by using sampled data generated with 10 \% and 20 \% variance for the actual energy information (\figref{fig:profit&distribution} (b)). The average value of each model result is \$9.400 $\times 10^7$, \$9.398 $\times 10^7$, and \$9.370 $\times 10^7$, and even if the accuracy of the prediction value is low, the profit is still 99.98 \% and 99.68 \% compared to when the prediction is perfect.

While the IP state model has the advantage that it does not need curtailed energy prediction model, the EP state model can show a higher performance than the IP state model if the prediction model can predict the pattern flow of the incoming curtailed energy well even though it is difficult to predict an accurate value.

\subsection{Qualitative analysis of the trained DRL policy}\label{sec:full}
\begin{figure}
    \includegraphics[width=\textwidth]{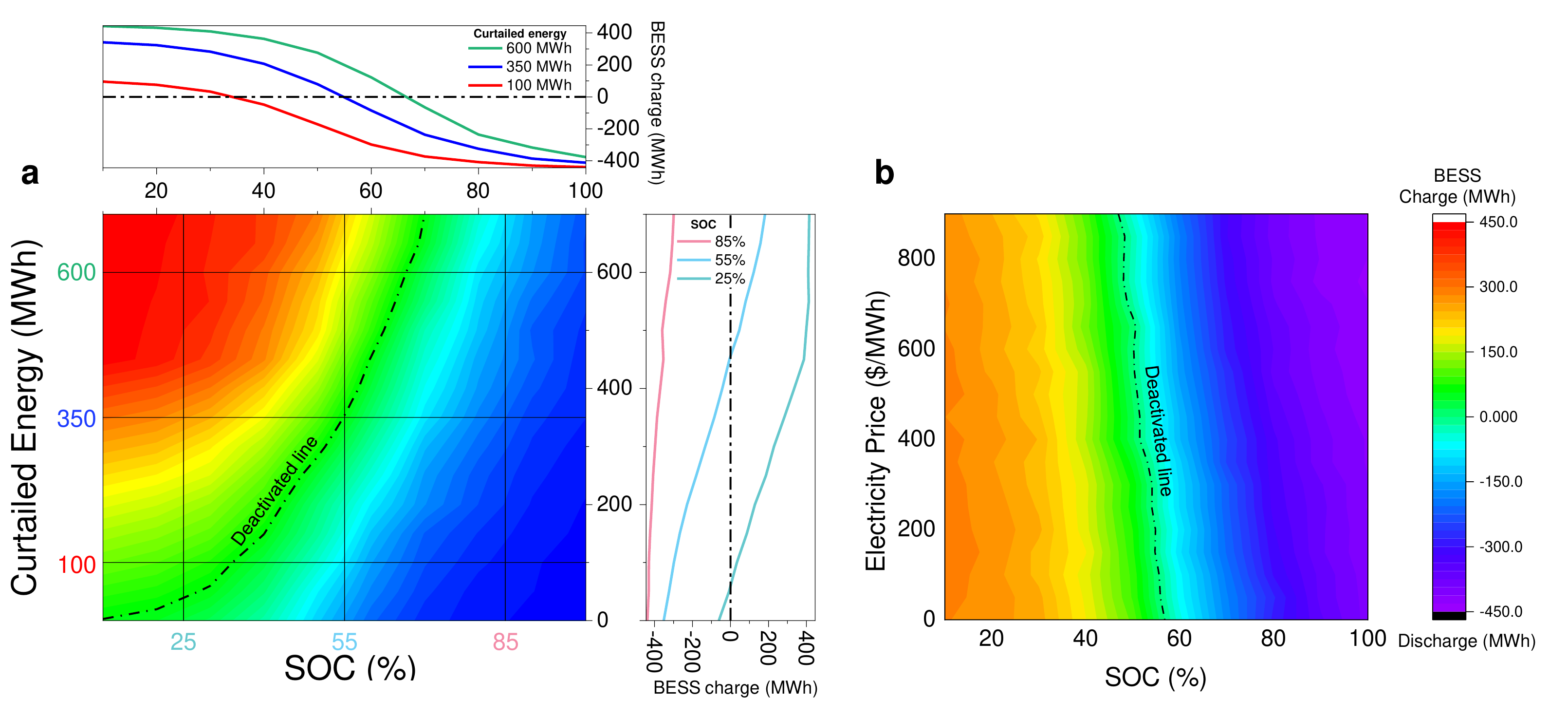}
    \caption{Action mapping of the BESS. (a) BESS action tendency according to curtailed energy and SOC. (b) BESS action tendency according to electricity price and SOC. The dashed line indicates the point where BESS is deactivated (no charging and no discharging).}
    \label{fig:contour_ESS}
\end{figure}

\begin{figure}
    \includegraphics[width=\textwidth]{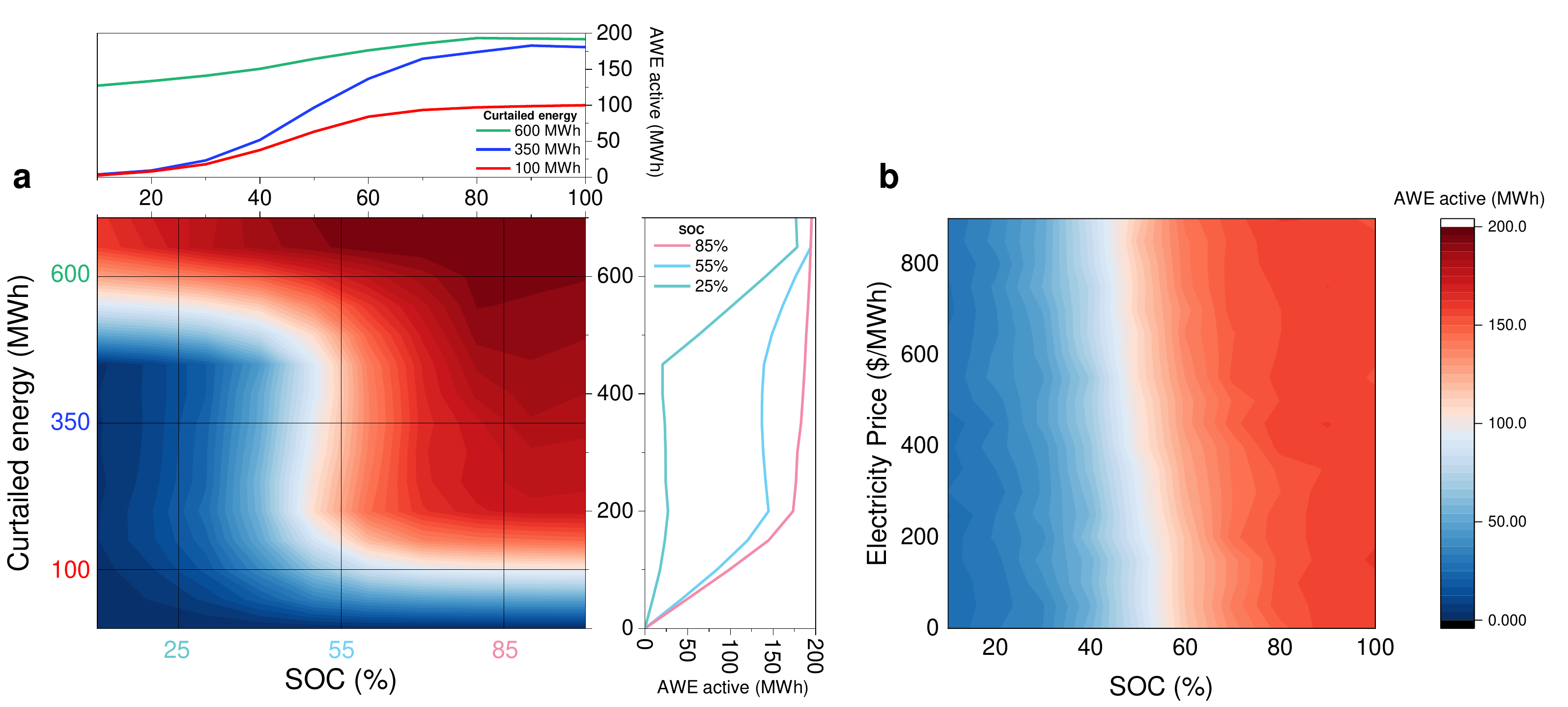}
    \caption{Action mapping of the AWE. (a) AWE action tendency according to curtailed energy and SOC. (b) AWE action tendency according to electricity price and SOC.}
    \label{fig:contour_AWE}
\end{figure}

To find out how the state affects the agent's action, we averaged and visualized the actions taken by the agent in a particular state. Two different situations were analyzed: first, the electricity price is determined randomly within a certain range, and 2000 samples are prepared for the average action value according to a given curtailed energy and SOC (\figref{fig:contour_ESS} (a) and \figref{fig:contour_AWE} (a)). Second, the curtailed energy is randomly determined and shows the effect of electricity price and SOC for the actions (\figref{fig:contour_ESS} (b) and \figref{fig:contour_AWE} (b)).

For a higher amount of curtailed energy and lower SOC, the size of the charge power shows an increasing tendency (\figref{fig:contour_ESS} (a)). It is expected that the action is going to charge when there are enough curtailed energy and storage. Moreover, a larger amount of curtailed energy allows a wider range of action choices. The right side contour profile for the SOC shows that as the SOC becomes lower and the amount of curtailed energy increases, the rate of charging gradually increases linearly. However, at some point, it appears as a broken line that can be seen on the 25 \% line. This is because when the SOC is below a certain level and a sufficient amount of curtailed energy is entered, the action determines charging to the maximum power capacity of the BESS, irrespective of the electricity price. Likewise, we can see the logic of this change of rate and specific points without directly calculating it through the action and state mapping.

In the case of AWE, as the SOC becomes higher and the curtailed energy increases, the hydrogen production increases (\figref{fig:contour_AWE} (a)). The contour line of the zero value is observed almost in the form of a step. When the SOC is less than a half, the AWE is hardly activated until the energy is near 450 MWh. This means that since charging the BESS is more valuable than hydrogen production, the agent tries to charge to the maximum amount of BESS power capacity, i.e., 450 MWh. Then, when more than 450 MWh of energy is received, the difference in the energy is used to activate the AWE. The right side of the contour indicates that as more curtailed energy is received, the hydrogen production increases because BESS has a tendency to discharge at this situation. However, when the energy is less than the maximum capacity of the AWE, the AWE is activated in proportion to the incoming curtailed energy.

For the electricity price and SOC case, neither the BESS nor the AWE show a significant dependence on the electricity price; however, a large dependence on the SOC is observed. The two variables show similar tendencies: expensive electricity price and high value of SOC increase discharging (\figref{fig:contour_ESS} (b)) and the production rate of hydrogen (\figref{fig:contour_AWE} (b)). For the BESS, the expensive electricity price makes is more beneficial when discharging the energy. Thus, even if the SOC value is low, there is a tendency to discharge a larger amount. AWE uses the curtailed energy alone when the BESS is discharged. Thus, the amount of curtailed energy for AWE activation increases and the production of hydrogen also increases. 
Another feature of this contour graphs is that, extreme values are not displayed. The effect of the electricity price for the BESS and AWE is less significant than that of the curtailed energy. Therefore, the randomly given curtailed energy value becomes a significant factor. Since averaged action values are drawn in these contour graphs, there are no extreme values of action here.

\begin{figure}
	\includegraphics[width=\textwidth]{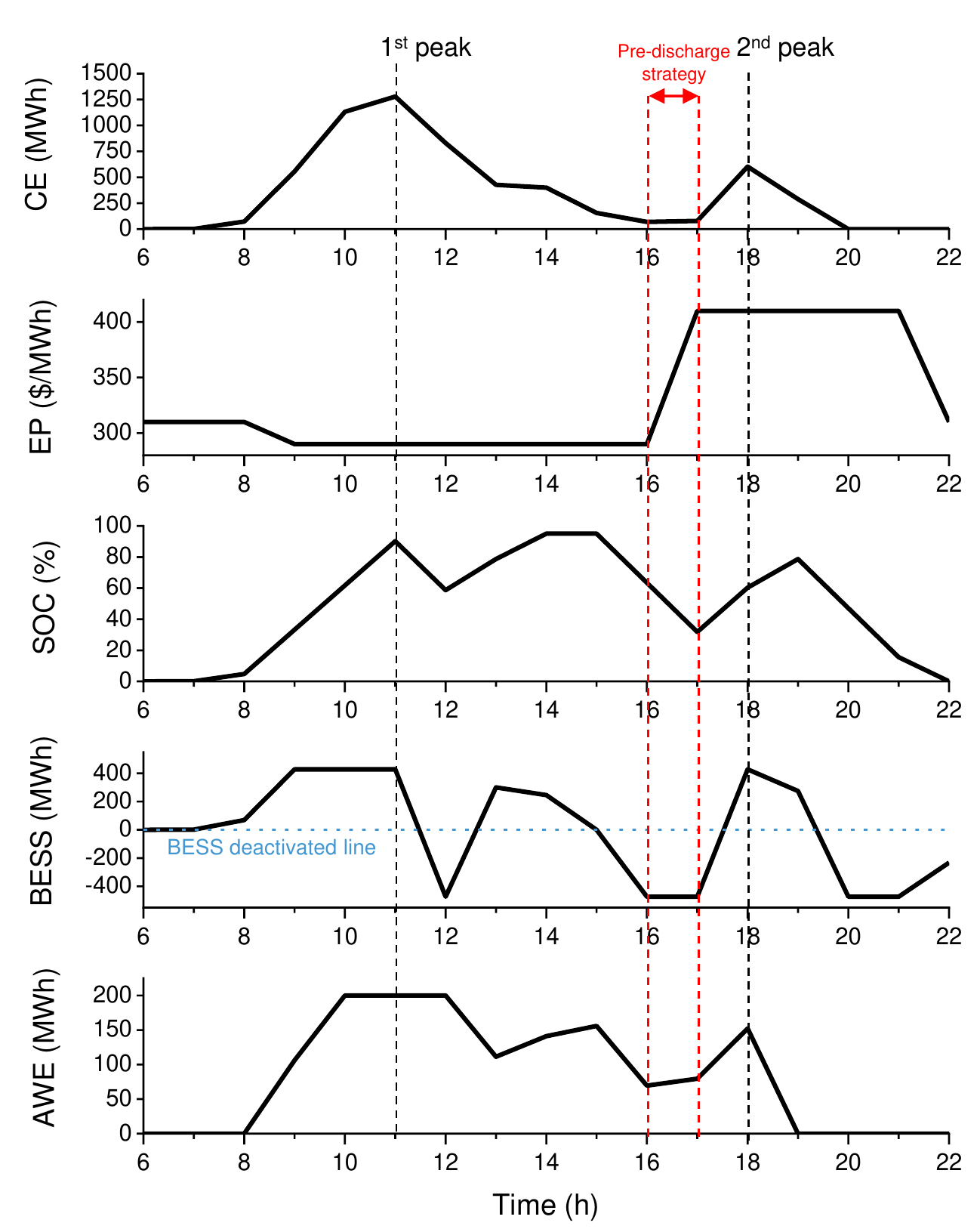}
	\caption{A specific-day example showing the action taken by the agent according to the amount of incoming energy during the day.}
	\label{fig:action_example}
\end{figure}

To see what actions were taken specifically at each step based on the current state, an illustrative example to demonstrate the action of our DRL model for one specific day, which is a part of the evaluation results in 2021 data, is shown in \figref{fig:action_example}. On this day, there were two major peaks in the curtailed energy input, even though usually only one peak occurs. In the first peak, the action first appears in the form of charging electricity, and then appears in the form of activating AWE when the amount of curtailed energy increases over the BESS power, because it is more valuable to store the energy in the BESS. Then, just before the SOC is full, the agent made a decision to discharge to avoid penalty, and the hydrogen production activation is maximally activated during discharging. It can be said that it is a general decision until charging is performed again after lowering the SOC or wait for a while until the electricity price rises at the end of the peak. However, when the second energy peak approaches, a different pattern of action is observed. Looking at the behavior at time 16 h in the SOC graph, it is discharged even though the SOC is not fully charged. This is because the agent decided that it would be advantageous to discharge in advance before the SOC would be full while charging at the second peak. Additionally, at time 17 h, the electricity price is higher than before; thus, the agent delays the sale of electricity at time 15 h to get more profit.

\section{Conclusions}\label{sec:con}
In this study, we developed an optimal decision-making artificial intelligence for hybrid energy storage systems, based on DRL methods. It shows a higher performance compared to SO under the curtailed renewable energy uncertainty, and achieves an optimal management. We quantitatively analyzed the performance of the DRL agent with deterministic MILP and SO results. The corresponding results guarantee about 90 \% profit, compared to what was solved with the deterministic MILP, and larger profits than those obtained from the SO. Furthermore, while the solutions of the SO depend on a limited number of scenarios, the proposed DRL agent naturally relies on stochastic policy networks that yield even outperformed solutions under uncertainties by covering diverse scenario. Interestingly, the action mapping analysis showed that the action decided by the DRL agent follows human intuitions, implying that the DRL agent not just memorizes the optimal policy at specific states but also learns the logical decision criteria depending on the curtailed energy, electricity price, and SOC. The results obtained in this study suggest the possibility of DRL could be used even in more complex systems, where multiple sources are used, and the real-time decision making in energy management systems.

\section*{Data Availability}
Datasets related to this article can be found at http://www.caiso.com/informed/Pages/ManagingOversupply.aspx, hosted at California Independent System Operator (CAISO) \cite{CAISO_data}.

\section*{Code Availability}
The python code and the trained neural networks used in this work are available under MIT licence in the GitHub repository (https://github.com/kangdj6358/EMSRL).

\section*{Author Contribution}
D.K.and D.K.: Conceptualization, Methodology, Software, Validation, Writing - Original Draft. S.H.: Resources, Writing - Review \& Editing. H.N.: Software, Data Curation. W.B.L.: Supervision. J.J.L.: Conceptualization, Writing - Review \& Editing, Supervision, Project administration, Funding acquisition. J.N.: Conceptualization, Methodology, Writing - Review \& Editing, Supervision, Project administration, Funding acquisition.

\section*{Declaration of Competing Interest}
The authors declare no competing interests.

\section*{Acknowledgements}
This work was supported by a National Research Foundation of Korea (NRF) grant funded by the Korean government (2021R1A4A3025742).

\bibliography{reference}

\begin{thebibliography}{10}
\expandafter\ifx\csname url\endcsname\relax
  \def\url#1{\texttt{#1}}\fi
\expandafter\ifx\csname urlprefix\endcsname\relax\def\urlprefix{URL }\fi
\expandafter\ifx\csname href\endcsname\relax
  \def\href#1#2{#2} \def\path#1{#1}\fi

\bibitem{DINCER2000157}
I.~Dincer,
  \href{https://www.sciencedirect.com/science/article/pii/S1364032199000118}{Renewable
  energy and sustainable development: a crucial review}, Renewable and
  Sustainable Energy Reviews 4~(2) (2000) 157--175.
\newblock \href
  {http://dx.doi.org/https://doi.org/10.1016/S1364-0321(99)00011-8}
  {\path{doi:https://doi.org/10.1016/S1364-0321(99)00011-8}}.

\bibitem{GOLDEN201536}
R.~Golden, B.~Paulos,
  \href{https://www.sciencedirect.com/science/article/pii/S1040619015001372}{Curtailment
  of renewable energy in california and beyond}, The Electricity Journal 28~(6)
  (2015) 36--50.
\newblock \href {http://dx.doi.org/https://doi.org/10.1016/j.tej.2015.06.008}
  {\path{doi:https://doi.org/10.1016/j.tej.2015.06.008}}.

\bibitem{IMPRAM2020100539}
S.~Impram, S.~{Varbak Nese}, B.~Oral,
  \href{https://www.sciencedirect.com/science/article/pii/S2211467X20300924}{Challenges
  of renewable energy penetration on power system flexibility: A survey},
  Energy Strategy Reviews 31 (2020) 100539.
\newblock \href {http://dx.doi.org/https://doi.org/10.1016/j.esr.2020.100539}
  {\path{doi:https://doi.org/10.1016/j.esr.2020.100539}}.

\bibitem{9224611}
M.~S. Alam, F.~S. Al-Ismail, A.~Salem, M.~A. Abido, High-level penetration of
  renewable energy sources into grid utility: Challenges and solutions, IEEE
  Access 8 (2020) 190277--190299.
\newblock \href {http://dx.doi.org/10.1109/ACCESS.2020.3031481}
  {\path{doi:10.1109/ACCESS.2020.3031481}}.

\bibitem{BIRD2016577}
L.~Bird, D.~Lew, M.~Milligan, E.~M. Carlini, A.~Estanqueiro, D.~Flynn,
  E.~Gomez-Lazaro, H.~Holttinen, N.~Menemenlis, A.~Orths, P.~B. Eriksen, J.~C.
  Smith, L.~Soder, P.~Sorensen, A.~Altiparmakis, Y.~Yasuda, J.~Miller,
  \href{https://www.sciencedirect.com/science/article/pii/S1364032116303161}{Wind
  and solar energy curtailment: A review of international experience},
  Renewable and Sustainable Energy Reviews 65 (2016) 577--586.
\newblock \href {http://dx.doi.org/https://doi.org/10.1016/j.rser.2016.06.082}
  {\path{doi:https://doi.org/10.1016/j.rser.2016.06.082}}.

\bibitem{SABER2019414}
H.~Saber, M.~Moeini-Aghtaie, M.~Ehsan, M.~Fotuhi-Firuzabad,
  \href{https://www.sciencedirect.com/science/article/pii/S014206151731092X}{A
  scenario-based planning framework for energy storage systems with the main
  goal of mitigating wind curtailment issue}, International Journal of
  Electrical Power \& Energy Systems 104 (2019) 414--422.
\newblock \href
  {http://dx.doi.org/https://doi.org/10.1016/j.ijepes.2018.07.020}
  {\path{doi:https://doi.org/10.1016/j.ijepes.2018.07.020}}.

\bibitem{8125737}
X.~Dui, G.~Zhu, L.~Yao, Two-stage optimization of battery energy storage
  capacity to decrease wind power curtailment in grid-connected wind farms,
  IEEE Transactions on Power Systems 33~(3) (2018) 3296--3305.
\newblock \href {http://dx.doi.org/10.1109/TPWRS.2017.2779134}
  {\path{doi:10.1109/TPWRS.2017.2779134}}.

\bibitem{williams2013model}
H.~P. Williams, Model building in mathematical programming, John Wiley \& Sons,
  2013.

\bibitem{LI20151067}
C.~Li, H.~Shi, Y.~Cao, J.~Wang, Y.~Kuang, Y.~Tan, J.~Wei,
  \href{https://www.sciencedirect.com/science/article/pii/S1364032114007825}{Comprehensive
  review of renewable energy curtailment and avoidance: A specific example in
  china}, Renewable and Sustainable Energy Reviews 41 (2015) 1067--1079.
\newblock \href {http://dx.doi.org/https://doi.org/10.1016/j.rser.2014.09.009}
  {\path{doi:https://doi.org/10.1016/j.rser.2014.09.009}}.

\bibitem{6936947}
L.~S. Vargas, G.~Bustos-Turu, F.~Larraín, Wind power curtailment and energy
  storage in transmission congestion management considering power plants ramp
  rates, IEEE Transactions on Power Systems 30~(5) (2015) 2498--2506.
\newblock \href {http://dx.doi.org/10.1109/TPWRS.2014.2362922}
  {\path{doi:10.1109/TPWRS.2014.2362922}}.

\bibitem{6345450}
P.~Denholm, Energy storage to reduce renewable energy curtailment, in: 2012
  IEEE Power and Energy Society General Meeting, 2012, pp. 1--4.
\newblock \href {http://dx.doi.org/10.1109/PESGM.2012.6345450}
  {\path{doi:10.1109/PESGM.2012.6345450}}.

\bibitem{8769895}
Y.~Du, F.~Li, Intelligent multi-microgrid energy management based on deep
  neural network and model-free reinforcement learning, IEEE Transactions on
  Smart Grid 11~(2) (2020) 1066--1076.
\newblock \href {http://dx.doi.org/10.1109/TSG.2019.2930299}
  {\path{doi:10.1109/TSG.2019.2930299}}.

\bibitem{DREHER2022115401}
A.~Dreher, T.~Bexten, T.~Sieker, M.~Lehna, J.~Schütt, C.~Scholz, M.~Wirsum,
  \href{https://www.sciencedirect.com/science/article/pii/S0196890422001972}{Ai
  agents envisioning the future: Forecast-based operation of renewable energy
  storage systems using hydrogen with deep reinforcement learning}, Energy
  Conversion and Management 258 (2022) 115401.
\newblock \href
  {http://dx.doi.org/https://doi.org/10.1016/j.enconman.2022.115401}
  {\path{doi:https://doi.org/10.1016/j.enconman.2022.115401}}.

\bibitem{ZHANG2019112199}
B.~Zhang, W.~Hu, D.~Cao, Q.~Huang, Z.~Chen, F.~Blaabjerg,
  \href{https://www.sciencedirect.com/science/article/pii/S0196890419312051}{Deep
  reinforcement learning–based approach for optimizing energy conversion in
  integrated electrical and heating system with renewable energy}, Energy
  Conversion and Management 202 (2019) 112199.
\newblock \href
  {http://dx.doi.org/https://doi.org/10.1016/j.enconman.2019.112199}
  {\path{doi:https://doi.org/10.1016/j.enconman.2019.112199}}.

\bibitem{sutton2018reinforcement}
R.~S. Sutton, A.~G. Barto, Reinforcement learning: An introduction, MIT press,
  2018.

\bibitem{ZHANG2021113608}
G.~Zhang, W.~Hu, D.~Cao, W.~Liu, R.~Huang, Q.~Huang, Z.~Chen, F.~Blaabjerg,
  \href{https://www.sciencedirect.com/science/article/pii/S0196890420311365}{Data-driven
  optimal energy management for a wind-solar-diesel-battery-reverse osmosis
  hybrid energy system using a deep reinforcement learning approach}, Energy
  Conversion and Management 227 (2021) 113608.
\newblock \href
  {http://dx.doi.org/https://doi.org/10.1016/j.enconman.2020.113608}
  {\path{doi:https://doi.org/10.1016/j.enconman.2020.113608}}.

\bibitem{ZHANG2021114381}
B.~Zhang, W.~Hu, D.~Cao, T.~Li, Z.~Zhang, Z.~Chen, F.~Blaabjerg,
  \href{https://www.sciencedirect.com/science/article/pii/S0196890421005574}{Soft
  actor-critic –based multi-objective optimized energy conversion and
  management strategy for integrated energy systems with renewable energy},
  Energy Conversion and Management 243 (2021) 114381.
\newblock \href
  {http://dx.doi.org/https://doi.org/10.1016/j.enconman.2021.114381}
  {\path{doi:https://doi.org/10.1016/j.enconman.2021.114381}}.

\bibitem{pmlr-v32-silver14}
D.~Silver, G.~Lever, N.~Heess, T.~Degris, D.~Wierstra, M.~Riedmiller,
  \href{https://proceedings.mlr.press/v32/silver14.html}{Deterministic policy
  gradient algorithms}, in: E.~P. Xing, T.~Jebara (Eds.), Proceedings of the
  31st International Conference on Machine Learning, Vol.~32 of Proceedings of
  Machine Learning Research, PMLR, Bejing, China, 2014, pp. 387--395.

\bibitem{8103164}
K.~Arulkumaran, M.~P. Deisenroth, M.~Brundage, A.~A. Bharath, Deep
  reinforcement learning: A brief survey, IEEE Signal Processing Magazine
  34~(6) (2017) 26--38.
\newblock \href {http://dx.doi.org/10.1109/MSP.2017.2743240}
  {\path{doi:10.1109/MSP.2017.2743240}}.

\bibitem{en12122291}
Y.~Ji, J.~Wang, J.~Xu, X.~Fang, H.~Zhang,
  \href{https://www.mdpi.com/1996-1073/12/12/2291}{Real-time energy management
  of a microgrid using deep reinforcement learning}, Energies 12~(12).
\newblock \href {http://dx.doi.org/10.3390/en12122291}
  {\path{doi:10.3390/en12122291}}.

\bibitem{8331897}
E.~Foruzan, L.-K. Soh, S.~Asgarpoor, Reinforcement learning approach for
  optimal distributed energy management in a microgrid, IEEE Transactions on
  Power Systems 33~(5) (2018) 5749--5758.
\newblock \href {http://dx.doi.org/10.1109/TPWRS.2018.2823641}
  {\path{doi:10.1109/TPWRS.2018.2823641}}.

\bibitem{HUA2019598}
H.~Hua, Y.~Qin, C.~Hao, J.~Cao,
  \href{https://www.sciencedirect.com/science/article/pii/S0306261919301746}{Optimal
  energy management strategies for energy internet via deep reinforcement
  learning approach}, Applied Energy 239 (2019) 598--609.
\newblock \href
  {http://dx.doi.org/https://doi.org/10.1016/j.apenergy.2019.01.145}
  {\path{doi:https://doi.org/10.1016/j.apenergy.2019.01.145}}.

\bibitem{doi:10.1080/17445760.2014.974600}
M.~Rayati, A.~Sheikhi, A.~M. Ranjbar,
  \href{https://doi.org/10.1080/17445760.2014.974600}{Optimising operational
  cost of a smart energy hub, the reinforcement learning approach},
  International Journal of Parallel, Emergent and Distributed Systems 30~(4)
  (2015) 325--341.
\newblock \href
  {http://arxiv.org/abs/https://doi.org/10.1080/17445760.2014.974600}
  {\path{arXiv:https://doi.org/10.1080/17445760.2014.974600}}, \href
  {http://dx.doi.org/10.1080/17445760.2014.974600}
  {\path{doi:10.1080/17445760.2014.974600}}.

\bibitem{8742669}
V.-H. Bui, A.~Hussain, H.-M. Kim, Double deep <inline-formula> <tex-math
  notation="latex">$q$ </tex-math></inline-formula>-learning-based distributed
  operation of battery energy storage system considering uncertainties, IEEE
  Transactions on Smart Grid 11~(1) (2020) 457--469.
\newblock \href {http://dx.doi.org/10.1109/TSG.2019.2924025}
  {\path{doi:10.1109/TSG.2019.2924025}}.

\bibitem{ZHANG2020113063}
B.~Zhang, W.~Hu, J.~Li, D.~Cao, R.~Huang, Q.~Huang, Z.~Chen, F.~Blaabjerg,
  \href{https://www.sciencedirect.com/science/article/pii/S0196890420306075}{Dynamic
  energy conversion and management strategy for an integrated electricity and
  natural gas system with renewable energy: Deep reinforcement learning
  approach}, Energy Conversion and Management 220 (2020) 113063.
\newblock \href
  {http://dx.doi.org/https://doi.org/10.1016/j.enconman.2020.113063}
  {\path{doi:https://doi.org/10.1016/j.enconman.2020.113063}}.

\bibitem{CAISO_data}
Caiso curtailed energy dataset,
  \url{https://www.caiso.com/informed/Pages/ManagingOversupply.aspx}.

\bibitem{SHAMS2021}
M.~H. Shams, H.~Niaz, J.~Na, A.~Anvari-Moghaddam, J.~J. Liu,
  \href{https://www.sciencedirect.com/science/article/pii/S2352152X21007210}{Machine
  learning-based utilization of renewable power curtailments under uncertainty
  by planning of hydrogen systems and battery storages}, Journal of Energy
  Storage 41 (2021) 103010.
\newblock \href {http://dx.doi.org/https://doi.org/10.1016/j.est.2021.103010}
  {\path{doi:https://doi.org/10.1016/j.est.2021.103010}}.

\bibitem{kim2020active}
J.-J. Kim, B.-S. Chae, Y.-K. Lee, K.-H. Cho, An active battery charge
  management scheme with predicting power generation in ess, Smart Media
  Journal 9~(1) (2020) 84--91.

\bibitem{Electricity_price}
California time-of-use electricity price,
  \url{https://www.sce.com/residential/rates/Time-Of-Use-Residential-Rate-Plans}.

\bibitem{hydrogen_cost}
2021 green hydrogen price,
  \url{https://www.iea.org/reports/global-hydrogen-review-2021}.

\bibitem{moritz2018ray}
P.~Moritz, R.~Nishihara, S.~Wang, A.~Tumanov, R.~Liaw, E.~Liang, M.~Elibol,
  Z.~Yang, W.~Paul, M.~I. Jordan, et~al., Ray: A distributed framework for
  emerging $\{$AI$\}$ applications, in: 13th USENIX Symposium on Operating
  Systems Design and Implementation (OSDI 18), 2018, pp. 561--577.

\bibitem{RAY_RLlib}
Ray rllib document, \url{https://docs.ray.io/en/latest/rllib/index.html}.

\bibitem{https://doi.org/10.48550/arxiv.1707.06347}
J.~Schulman, F.~Wolski, P.~Dhariwal, A.~Radford, O.~Klimov,
  \href{https://arxiv.org/abs/1707.06347}{Proximal policy optimization
  algorithms} (2017).
\newblock \href {http://dx.doi.org/10.48550/ARXIV.1707.06347}
  {\path{doi:10.48550/ARXIV.1707.06347}}.

\bibitem{HyperOpt}
J.~Bergstra, D.~Yamins, D.~Cox, Making a science of model search:
  Hyperparameter optimization in hundreds of dimensions for vision
  architectures, in: International conference on machine learning, PMLR, 2013,
  pp. 115--123.

\end{thebibliography}

\end{document}


\captionsetup[figure]{labelfont={bf},labelformat={default},labelsep=period,name={Fig.}}
\captionsetup[table]{labelfont={bf},labelformat={default},labelsep=newline,name={Table}, justification=raggedright, singlelinecheck=off}
\maketitle
\tableofcontents
\clearpage

\newpage

\section{Reward design}
\begin{figure}
	\includegraphics[width=\textwidth]{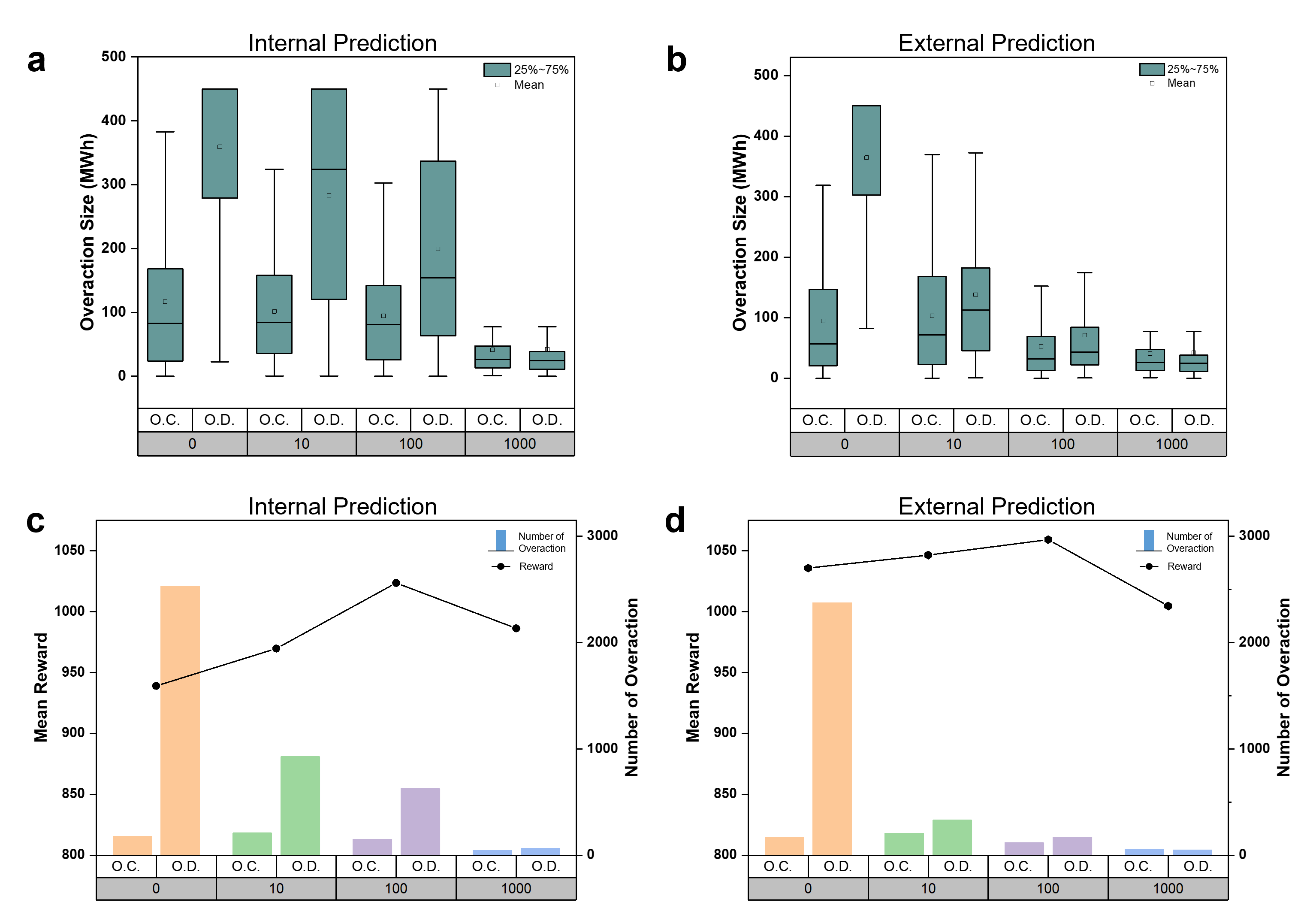}
	\caption{Comparison of the training results according to penalty. (a) Distribution of overaction sizes of IP type state. (b) Distribution of overaction sizes of EP type state. (c) Return and number of overaction for IP type state. (d) Return and number of overaction for EP type state.}
	\label{fig:penalty}
\end{figure}

Reward is one of the most important factor in the reinforcement learning and it can vary depending on the pursed direction of agent's character. Hence, we tried to adjust penalty scale with penalty scaling factor ($\mathcal{F}$). For both type of state, IP and EP, the number of overaction was gradually decreased while the penalty scaling factor ($\mathcal{F}$) increased (\figref{fig:penalty}). When the penalty scaling factor set in low, the agent learned greedy method that makes excessively large action, so that the BESS system could have been damaged. Otherwise, when the penalty was set in high, the agent acted passively not to exceed the curtailed energy. However, the mean reward seems to increase as the penalty scaling factor increases, but it shows a form of decreasing again at some point. At first, the penalty makes the agent learns to use the most of the curtailed energy avoid the large penalty value and it leads to rise in profits. Later, the excessively conservative action leads to a decrease in profits. Therefore, we used the penalty scaling factor with an proper interval between high reward and low overaction which is 100, here.

\begin{equation}
  \mathcal{P}_t =
  \begin{cases}
    P^{CH}_{t}-(100-SOC_{t})\times BESS^{cap} & \text{if $P^{CH}_{t} \geq (100-SOC_{t}) \times BESS^{cap} $} \\
    P^{D}_{t}-(SOC_{t}-SOC_{min})\times BESS^{cap} & \text{if $P^{D}_{t} \geq (SOC_{t}-SOC_{min})\times BESS^{cap}$} \\
    0 & \text{otherwise}
  \end{cases}
\end{equation}

\begin{eqnarray}
\mathcal{R}_t = REV_{t}-C^{VOM}_{t}-(C^{FO}+C^{ACC})/T-\mathcal{P}_t \times \mathcal{F} \label{eq:reward_penalty}
\end{eqnarray}

\section{Training under uncertainties}\label{sec:full}

The strengths of RL appear manifestly when the environment has uncertainties and is evident as uncertainties increases. The optimal solution of mathematical programming is not very liquid on the changes of the environment, but RL is less affected by the changes of the environment. To obtain the solution under uncertainty, stochastic optimization or robust optimization are the representative methods in the mathematical programming field. However, they take more conservative values as the uncertainty increases and computational cost and time cost increase exponentially. For solving stochastic optimization with this problem, GAMS software was used with the CPLEX solver. Unfortunately, the problem with two types of uncertainties (i.e. curtailment energy, production prices) was not solved due to the memory limitations of GAMS. To improve the accuracy of the stochastic optimization result, the number of scenarios should be increased and reduce the primal-dual optimality gap, but it is close to impossible without enormous computer costs. On the other hand, training RL policy does have less impact on reflecting the uncertainty data. All cases that training under uncertainties did not take CPU memories and time as much as SO and show similar or better results compared to training without uncertainties while the SO solution gets worse with the size of uncertainties. The reason why this phenomenon appears conspicuously is that the length of the period is long in this problem. A long period and short time steps make the data size larger. These characteristics make RL stand out and are needed in multi-period problems.

\begin{figure}
	\includegraphics[width=\textwidth]{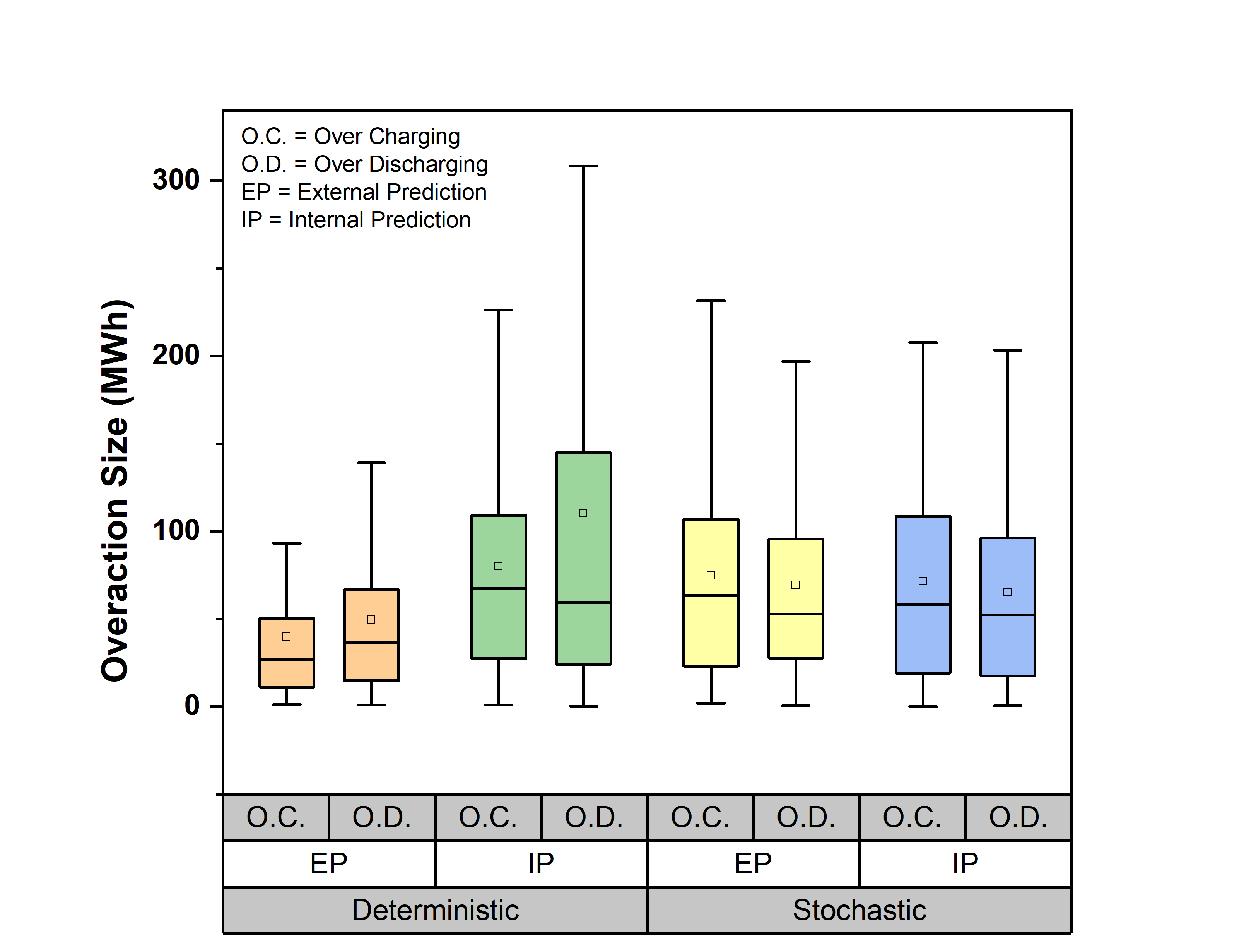}
	\caption{Comparison of the degree of over-action according to the state type.}
	\label{fig:overaction}
\end{figure}

When the two type of RL model, External Prediction (EP) and Internal Prediction (IP), are trained with stochastic data that contains uncertainties in curtailed energy, the results of reward and over-action is represented in Table \ref{tbl:uncertainty} and \figref{fig:overaction}. It shows that both models with uncertainty has lower scores and more frequent over-action. This meant that giving variability to the data to learn policies to prepare for uncertainty was not effective. Since the evaluation result was not significantly different from the training result, it can be seen that the high-reward model is not overfitted, and it is important to learn the overall data pattern rather than the data with variability at every step for good training. For this reason, the models in this paper, the models trained on deterministic data are used.

\begin{table}[t]
	\centering
	\caption{Comparison between four models about reward and over-action. Over-action value means the number of how many times the over-action happened in one episode (8784 hours). The value in parentheses represents the average value of exceeded action when over-action occurs once.}
	\label{tbl:uncertainty}
	\resizebox{\textwidth}{!}{\begin{tabular*}{1.5\textwidth}{@{\extracolsep{\fill}}llll}
		\hline
		Model         & Reward   & over-charge   &  over-discharge \\
		\hline
		EP$_{\rm det}$     &  1068      & 118(39)   & 86(49)\\
		IP$_{\rm det}$    & 988   &  156(79)  & 88(107)\\
		EP$_{\rm stc}$     &992& 127(74)    &  181(69)\\
		IP$_{\rm stc}$    & 894   &   97(70)  & 180(65)\\
		\hline
	\end{tabular*}}
\end{table}

For the reference, the number of overaction in Table \ref{tbl:uncertainty} is smaller than in Supplementary Note 1 because the penalty design was changed more delicately. The final RL model includes the penalty related to AWE as well as ESS. The penalty of ESS and AWE are summed into one penalty and reflected in the reward. If the penalty scaling factor is the power to reduce the final penalty, it seems to maintain the same level of penalty by reducing the ESS penalty more strongly when the AWE penalty is added.

\section{Length of period for training data}

\begin{figure}
	\includegraphics[width=\textwidth]{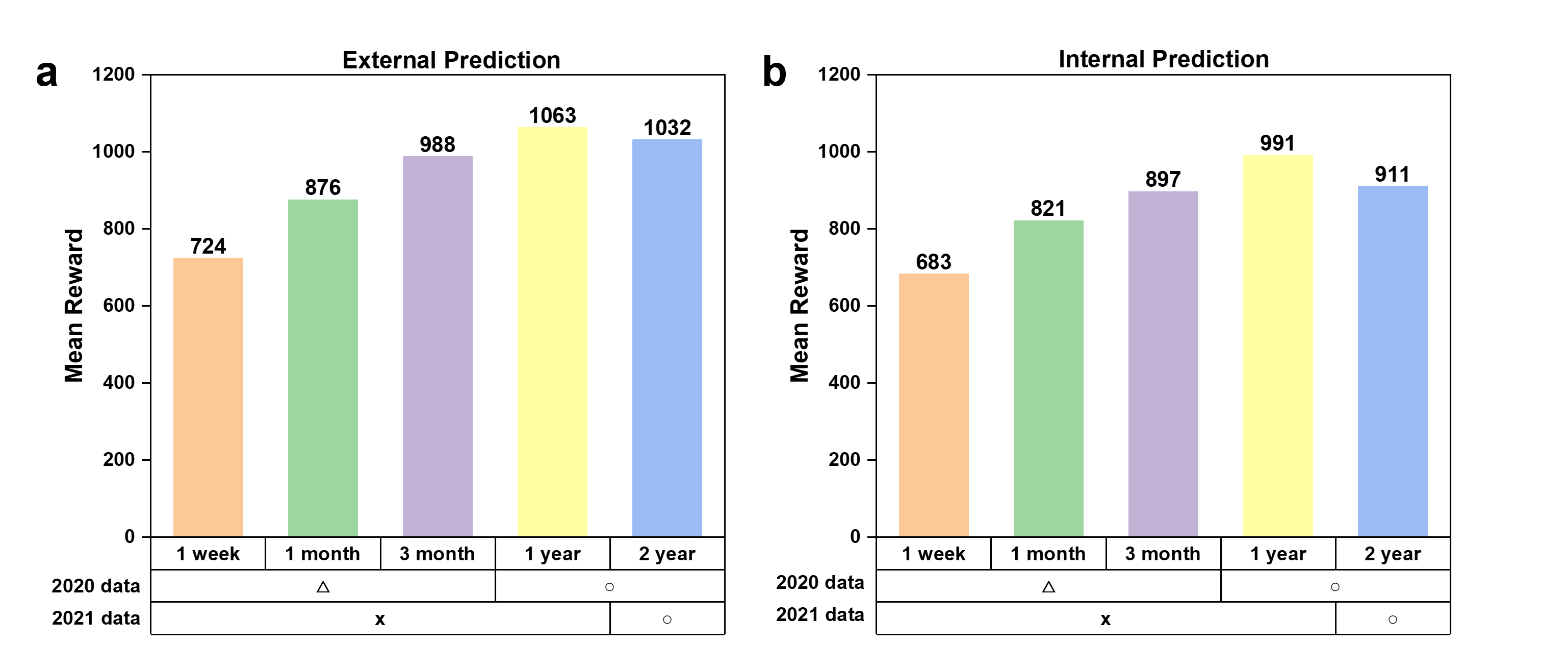}
	\caption{Comparison of RL training results according to training data period. The circle symbol ($\circ$) means that the all data for that year was used and the triangle symbol ($\vartriangle$) means that only part of the data for that year was used.}
	\label{fig:evaluation_results_training_period}
\end{figure}

In RL, the amount of trainable data is one of the most important factors to train well. RL can have higher accuracy if the more high-quality training data is added, but there are some restraints due to the limitation of data. We tried to adjust the period of the training data in order to use the existing data to the maximum for efficient training. To determine how much period of data to train, we evaluate some samples to change length of period for training data. \figref{fig:evaluation_results_training_period} shows the difference between the model trained with data for one week, one month, three months, one year and for two years. Through the graph, although even if there were not enough train data, RL could decide effective actions. As a result of the 2020 data training, training with 1 year of data had the best performance. When evaluated for 2019 and 2021, there was no significant difference between the policies trained in 1 year and 2 years. Therefore, we determined to use one year data for training.

\section{Composition of observable state}
\begin{figure}
	\includegraphics[width=\textwidth]{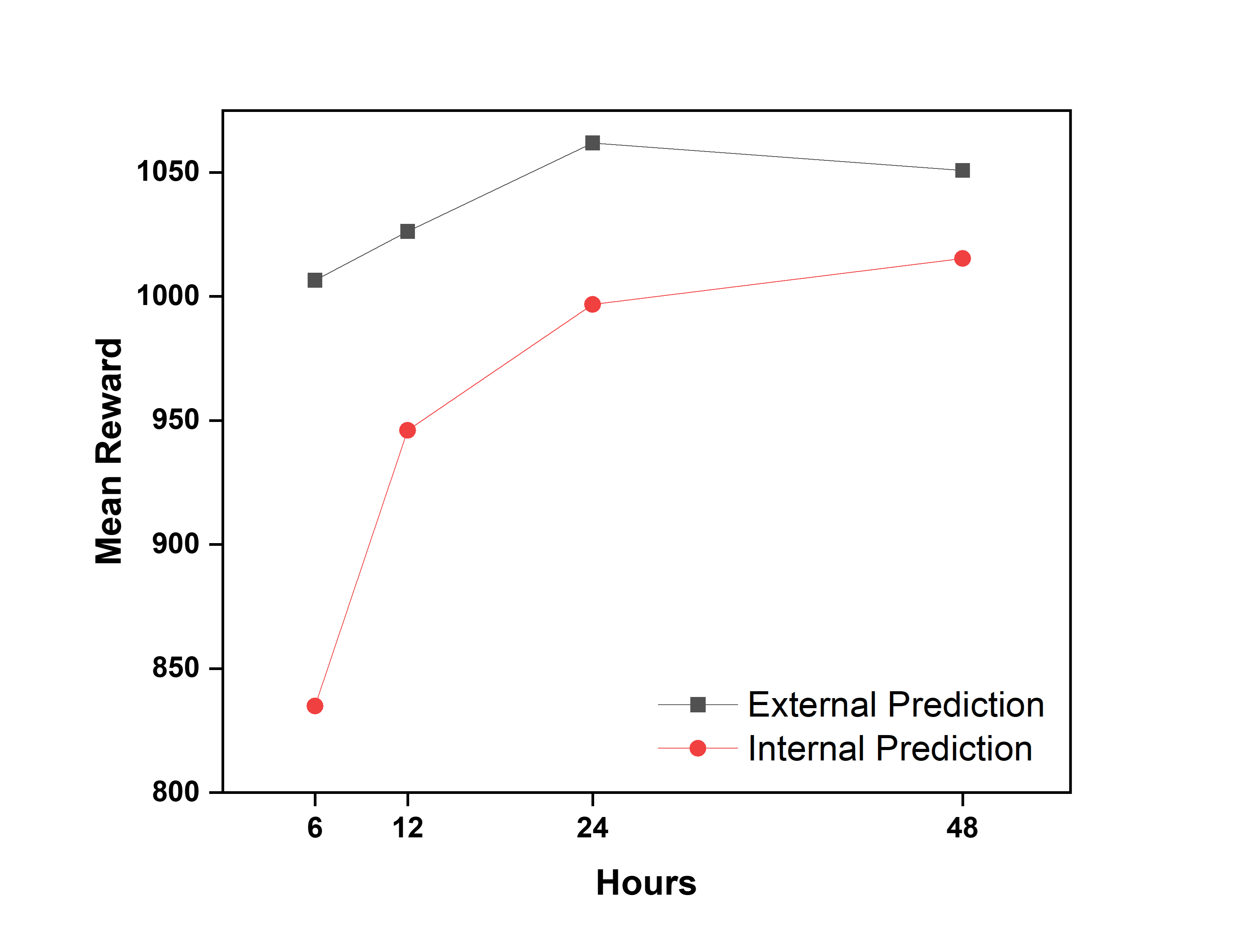}
	\caption{Comparison of RL training results according to observable state information.}
	\label{fig:evaluation_results_observable_state}
\end{figure}

In our RL model, the state contains the SOC, remaining BESS capacity, and the periodic data of electricity price, hydrogen price, and energy curtailed data. The agent act through the information of current state, and the amount of information effects the training result. To find a reasonable period of observable state value, we trained model while changing the period of data entering the state. \figref{fig:evaluation_results_observable_state} shows the evaluation results of 6, 12, 24, and 48 hour observation state trained models. In EP model, the period means how long an external prediction model can predict energy curtailed data. For example, 6 hour period EP model has 20 state information: the SOC, remaining BESS capacity, 6 hour data of electricity price, hydrogen price, amount of curtailed energy from the state time. For IP model, the period means how much data for the previous period to predict the curtailed energy of the current state. Likewise, 6 hour period IP model has 20 state information: the SOC, remaining BESS capacity, 6 hour data of electricity and hydrogen price, and 6 hour previous curtailment energy data that will be used to predict the current state. The result shows more information in the state could help the agent learning more effectively. It is obvious that if the agent has more information, it can decide more accurate action. Looking both EP and IP graph, the accuracy raised extremely through 6 to 24 hour, but similar with 24 and 48 hour. Due to these results, we set the period 24 hours as the observable time to avoid over-fitting.

\section{Economic model}
For the economic model of cost estimation for BESS and AWE consists of the annualized capital cost ($C^{ACC}$) and operation \& management cost ($C^{O\&M}$). The annualized capital cost can be calculated as follows,
\begin{eqnarray}
C^{ACC,BESS} = C^{BESS} \times \text{CRF} = C^{BESS} \times (\text{r})/((1+\text{r})^{n}-1) \label{eq:BESS capital cost} \\
C^{ACC,AWE} =  C^{AWE} \times \text{CRF} = C^{AWE} \times (\text{r})/((1+\text{r})^{n}-1) \label{eq:AWE capital cost}
\end{eqnarray}
where $C^{BESS}$ and $C^{AWE}$ are the cost of BESS and AWE, respectively; CRF is the capital recovery factor; r and n are the inflation rate and project life time, respectively. The life time of the all the system are assumed as 10 years.

The operation \& management cost for the BESS can be expressed as follows,
\begin{eqnarray}
C^{O\&M,BESS} = C^{VOM,BESS} + C^{FO,BESS} \label{eq:BESS O&Mcost} \\
C^{VOM,BESS} = \sum_{t=1}^{T}P^{DH}_{t} \times c^{VOM,BESS} \label{eq:BESS VOM cost} \\
C^{FO,BESS} = BESS^{CAP} \times c^{FO, BESS} \label{eq:BESS FO cost}
\end{eqnarray}
where $c^{VOM,BESS}$ is the variable operation \& management cost per MWh for BESS; $c^{FO, BESS}$ is the fixed cost per MWh for BESS.

For the AWE, the operation \& management cost can be expressed as follows,
\begin{eqnarray}
C^{O\&M,AWE} &=& C^{VOM,AWE} + C^{FO,AWE} \label{eq:AWE O&Mcost} \\
C^{VOM,AWE} &=& \sum_{t=1}^{T}H^{AWE}_{t} \times ((S^{H_{2}O} \times F^{H_{2}O}) + (S^{KOH} \times F^{KOH}) \\
&&+ (S^{steam} \times F^{steam}) + (S^{N_{2}} \times F^{N_{2}})) \nonumber \label{eq:AWE O&M cost} \\
C^{FO,AWE} &=& 0.05 \times AWE^{P,MAX} \times C^{AWE} \label{eq:AWE FO cost}
\end{eqnarray}
where $S^{H_{2}O}$, $S^{KOH}$, $S^{steam}$ and $S^{N_{2}}$ are the sale price of deionized water, KOH, steam and nitrogen, respectively; $F^{H_{2}O}$, $F^{steam}$, $F^{N_{2}}$ and $F^{N_{2}}$ are the flow of deionized water, KOH, steam and nitrogen per kg of hydrogen, respectively.

\bibliographystyle{unsrt}
\bibliography{ref}